\documentclass{article}
\usepackage{ijcai23}

\usepackage{times}
\usepackage{soul}
\usepackage{url}
\usepackage[hidelinks]{hyperref}
\usepackage[utf8]{inputenc}
\usepackage[small]{caption}
\usepackage{graphicx}
\usepackage{amsmath}
\usepackage{amsthm}
\usepackage{booktabs}
\usepackage{algorithm}
\usepackage[noend]{algorithmic}
\usepackage{xcolor}
\usepackage{tikz}
\usetikzlibrary{positioning}
\usetikzlibrary{arrows.meta}
\usepackage{mathrsfs}
\usepackage{mathtools}
\usepackage{amssymb}
\usepackage{multicol,caption}
\usepackage{subcaption}
\usepackage{stmaryrd}
\usepackage{tikz-cd}
\usepackage{array}

\tikzset{
mynode/.style={
  draw,
  circle,
  text width=0.6cm,
  minimum size=0.6cm,
  align=center,
  fill=mygray
  }
}

\tikzset{
encoden/.style={
  draw,
  circle,
  text width=0.6cm,
  minimum size=0.6cm,
  align=center
  }
}

\tikzset{
mytext/.style={
  text width=0.6cm,
  minimum size=0.6cm,
  align=center
  }
}

\tikzset{
mysimple/.style={
  draw,
  circle,
  fill=mygray
  }
}

\tikzset{
arr/.style = {-{Triangle[length=2.5mm, width=2mm]}}
}
\tikzset{
arr1/.style = {->,line width=1pt}
}

\newcommand{\EIC}[1]{E_{IC}(#1)}
\newcommand{\eIC}[1]{e_{IC}(#1)}
\newcommand{\EIIL}[1]{E_{IIL}(#1)}
\newcommand{\eIIL}[1]{e_{IIL}(#1)}
\newcommand{\EISIL}[1]{E_{ISIL}(#1)}
\newcommand{\eISIL}[1]{e_{ISIL}(#1)}
\newcommand{\EISC}[1]{E_{ISC}(#1)}
\newcommand{\eISC}[1]{e_{ISC}(#1)}

\newcommand{\EI}[1]{E_{I}(#1)}
\newcommand{\eI}[1]{e_{I}(#1)}

\newcommand{\absi}{\boldsymbol{\alpha}}
\newcommand{\absii}{\boldsymbol{\beta}}

\newcommand{\scm}{\mathcal{M}}
\newcommand{\scmi}{\mathcal{M'}}
\newcommand{\scmii}{\mathcal{M''}}

\newcommand{\mpinv}[1]{#1^{+}}

\newtheorem{example}{Example}

\newtheorem{definition}{Definition}
\newtheorem*{definition*}{Definition}
\newtheorem{proposition}{Proposition}
\newtheorem*{proposition*}{Proposition}

\title{Quantifying Consistency and Information Loss for Causal Abstraction Learning}

\author{Anonymous Submission}

\author{
Fabio Massimo Zennaro$^1$
\and
Paolo Turrini$^1$\And
Theodoros Damoulas$^1$
\affiliations
$^1$University of Warwick,Coventry, United Kingdom
\emails
\{fabio.zennaro, p.turrini, t.damoulas\}@warwick.ac.uk,
}

\begin{document}

\maketitle





\begin{abstract}
Structural causal models provide a formalism to express causal relations between variables of interest. Models and variables can represent a system at different levels of abstraction, whereby relations may be coarsened and refined according to the need of a modeller.
However, switching between different levels of abstraction requires evaluating a trade-off between the consistency and the information loss among different models.
In this paper we introduce a family of interventional measures that an agent may use to evaluate such a trade-off. We consider four measures suited for different tasks, analyze their properties, and propose algorithms to evaluate and learn causal abstractions. Finally, we illustrate the flexibility of our setup by empirically showing how different measures and algorithmic choices may lead to different abstractions.

\end{abstract}

\section{Introduction}

In his IJCAI 2022 keynote talk, Judea Pearl argued that reasoning with causality is among the biggest challenges of modern AI. \textit{Structural causal models} (SCM) \cite{pearl2009causality} were introduced to address this challenge as a rigorous graph-based formalism explicitly encoding causal relations between variables. Analyzing causes and effects, however, implicitly requires the assumption of a given \textit{level of abstraction} (LA) at which variables are observed. The same system may indeed be modelled at different LAs depending on the resolution a modeller or a decision-making agent are considering. Choosing the appropriate scale for modelling, analyzing and controlling a system is a fundamental challenge in science and decision-making, with instances ranging from ecological multi-scale modelling \cite{levin1992problem} to neural population coarse-graining \cite{schmutz2020mesoscopic}.



Within the context of causal models, evaluating which abstraction is the ``correct" one is a nontrivial challenge in itself. A few approaches have been proposed in the literature to express relationships of abstraction between SCMs 
\cite{rubenstein2017causal,beckers2018abstracting,rischel2020category}, with some of them offering quantitative ways to assess the degree of approximation (or error) introduced by an abstraction in terms of interventional consistency (IC) \cite{rischel2020category,rischel2021compositional}. However, understanding which LA is the optimal one requires balancing  potentially conflicting properties.
For instance, while assessing among different candidate abstractions, an agent may well be concerned with information loss when probability distributions over fine-grained random variables are compressed to fit coarser variables. Although abstractions are a key part of causal reasoning, we still lack a theoretical framework specifying flexible measures of approximation and how to use them to learn optimal abstractions. 
 




\paragraph{Contributions.}
In this paper we bridge the gap in the literature by introducing and analyzing measures of abstraction approximation that capture consistency and information loss in causal abstraction.
Concretely: 
(i) we define a family of interventional measures of abstraction approximation (of which IC is a particular case) and analyze their properties;
(ii) we introduce algorithms for evaluating and learning causal abstractions based on our properties;
(iii) we illustrate how our measures are sensitive to their parameters and how they can capture different aspects of an abstraction. All in all, we provide a grounded set of measures that can be used for abstraction learning according to the specific aims at hand.


\paragraph{Related Literature.}

Abstraction is a fundamental component for general intelligence \cite{mitchell2021abstraction} and an important strategy for managing complexity and allowing artificial agents to achieve superhuman performance in challenging games \cite{KroerS18,Sandholm15a}. Coarsening of Bayesian networks or learning structures with bounded complexity in order to improve computational efficiency and representation has been studied, for instance, in \cite{chang1990refinement,elidan2008learning}.

In causal reasoning, the problem of abstraction between SCMs was first introduced by \cite{rubenstein2017causal}; the original framework was then further developed by \cite{beckers2018abstracting,beckers2020approximate}, and it has recently found application for interpretability \cite{geiger2021causal}.
Our work builds on the results of \cite{rischel2020category}, who provides a framework for evaluating IC grounded in category theory \cite{spivak2014category}. The problem of learning abstractions within this framework has been practically studied in \cite{zennaro2023jointly} on synthetic and real-world data. Here, we generalise the IC approach across multiple dimensions and study it both analytically and algorithmically.
%


Work on evaluating abstraction naturally connects to work on causal representation learning \cite{chalupka2017causal} and work on defining and measuring emergence \cite{hoel2017map,eberhardt2022causal}, too; these deal with causal systems at different LAs without the formalism of SCMs.


\paragraph{Paper Structure}

In Sec. \ref{sec:preliminaries} we introduce mathematical preliminaries for our framework. In Sec. \ref{sec:measures} we provide measures for abstraction approximation, and in Sec. \ref{sec:properties} we study their properties. We rely on these properties to discuss how to learn abstractions in Sec. \ref{sec:evaluating}. Finally, we illustrate our contributions empirically in Sec. \ref{sec:evaluation} and conclude in Sec. \ref{sec:conclusion}.


\section{Preliminaries}\label{sec:preliminaries}
We define here SCMs, interventions, and abstractions.

\begin{definition}[SCM \cite{pearl2009causality}] \label{def:SCM}
A structural causal model  (SCM) $\scm$ is a tuple $\langle \mathcal{X}, \mathcal{U}, \mathcal{F}, P(\mathcal{U}) \rangle$ with an underlying directed acyclic graph (DAG) $\mathcal{G}_\scm$ where:
\begin{itemize}
	\item $\mathcal{X}$ is a finite set of $N$ endogenous random variables $X_i$; each variable $X_i$ is associated with a finite set $\scm[X_i]=\{x_1, ..., x_M\}$ of outcomes; sets of variables are associated with the Cartesian product of the sets.
	\item $\mathcal{U}$ is a finite set of exogenous random variables.
	\item $\mathcal{F}$ is a finite set of $N$ measurable structural functions $f_i$, one for each endogenous variable $X_i$; a structural function $f_i: \scm[{Pa}(X_i)] \times \scm[\mathcal{U}] \rightarrow \scm[X_i]$, where $Pa(X_i) \subseteq \mathcal{X}$ denotes parents, defines deterministically the value of the random variable $X_i$. 
	\item $P(\mathcal{U})$ is a distribution over the exogenous variables. 
\end{itemize}
\end{definition}

Following \cite{rischel2020category}, we assume we have a finite number of endogenous variables, each one with a finite domain. Our definition implies that we are working with semi-Markovian SCMs. Notice, also, that the DAG structure implies a partial ordering $X_i \prec X_j$ of the endogenous variables according to reachability. SCMs allow us to study causality via interventions:

\begin{definition}[Intervention \cite{pearl2009causality}]
Given a SCM $\scm$, a variable-value pair $(\mathbf{X},\mathbf{x})$ such that for each $X_i$ in the set $\mathbf{X} \subseteq \mathcal{X}$ there is a $x_i \in \scm[X_i]$ in the set $\mathbf{x}$, an intervention $\iota: do(\mathbf{X}=\mathbf{x})$ is an operator that generates a new SCM $\scm_{\iota}$ by replacing the structural functions $f_i$ with the constants $x_i$.
\end{definition}

Let us now consider a base (low-level) model $\scm$ and an abstracted (high-level) model $\scm'$ and define abstraction:

\begin{definition}[Abstraction \cite{rischel2020category}]
An abstraction $\boldsymbol{\alpha}$  from SCM $\scm$ to SCM $\scmi$ is a tuple $\langle R, a, \alpha_i \rangle$ where:
\begin{itemize}
	\item $R \subseteq \mathcal{X}$ defines a subset of relevant variables in $\scm$;
	\item $a: R \rightarrow \mathcal{X'}$ is a surjective function mapping relevant variables $R$ in $\scm$ to variables in $\scmi$;
	\item $\alpha_i: \scm[a^{-1}(X'_i)] \rightarrow \scmi[X'_i]$ is a collection of surjective functions, one for each variable in $\scmi$, mapping the outcomes of variable(s) $a^{-1}(X'_i)$ onto the outcomes of variable $X'_i$.
\end{itemize}
\end{definition}

\begin{figure}
	\centering
	\begin{subfigure}{.17\textwidth}
		\centering
		\begin{tikzpicture}[shorten >=1pt, auto, node distance=1cm, thick, scale=0.7, every node/.style={scale=0.7}]
			\tikzstyle{node_style} = [circle,draw=black]
			\node[node_style] (S) at (0,0) {Sm};
			\node[node_style] (T) at (1.5,0) {Tar};
			\node[node_style] (C) at (3,0) {LC};
			
			\draw[->]  (T) to (C);
			\draw[->]  (S) to (T);
		\end{tikzpicture}
		\caption{}
		\label{fig:SCM0}
	\end{subfigure}
	\begin{subfigure}{.12\textwidth}
		\centering
		\begin{tikzpicture}[shorten >=1pt, auto, node distance=1cm, thick, scale=0.65, every node/.style={scale=0.65}]
			\tikzstyle{node_style} = [circle,draw=black]
			\node[node_style,blue] (S) at (0,0) {Sm'};
			\node[node_style,blue] (C) at (1.5,0) {Hea'};
			
			\draw[->,blue]  (S) to (C);
		\end{tikzpicture}
		\caption{}
		\label{fig:SCM1}
	\end{subfigure}
	\begin{subfigure}{.12\textwidth}
		\centering
		\begin{tikzpicture}[shorten >=1pt, auto, node distance=1cm, thick, scale=0.65, every node/.style={scale=0.65}]
			\tikzstyle{node_style} = [circle,draw=black]
			\node[node_style,red] (S) at (0,0) {Env''};
                \node[node_style,red] (T) at (0,-1.5) {Gen''};
			\node[node_style,red] (C) at (1.5,-.75) {LC''};
			
			\draw[->,red]  (S) to (C);
                \draw[->,red]  (T) to (C);
		\end{tikzpicture}
		\caption{}
		\label{fig:SCM2}
	\end{subfigure}
        \begin{subfigure}{.05\textwidth}
		\centering
		\begin{tikzpicture}[shorten >=1pt, auto, node distance=1cm, thick, scale=0.8, every node/.style={scale=0.8}]
			\tikzstyle{node_style} = [circle,draw=black]
			\node[node_style] (S) at (0,0) {*};
		\end{tikzpicture}
		\caption{}
		\label{fig:SCMstar}
	\end{subfigure}
	\caption{DAGs of four SCMs modelling causal relationships between smoking (Sm), tar deposits (Tar), genetic factors (Gen), environmental factors (Env), health index (Hea), and lung cancer (LC).}
\end{figure}
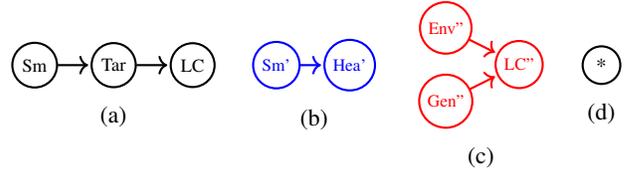

\begin{example}\label{ex:SCM0SCM1}
Consider two laboratories, Lab A and Lab B, having defined two models of lung cancer: the model in Fig. \ref{fig:SCM0} and the model in Fig. \ref{fig:SCM1}, respectively. A formal abstraction $\boldsymbol{\alpha}$ from the model of Lab A to the model of Lab B may be defined by choosing $R=\{Sm,LC\}$, $a:\{Sm \mapsto Sm', LC \mapsto Hea'\}$, and setting $\alpha_{Sm'},\alpha_{Hea'}$ to identities. A complete definition of the SCMs and the abstraction is provided in App. \ref{app:samplemodels}. 
\end{example}


A first measure to assess quantitatively an abstraction was suggested in \cite{rischel2020category} in the form of IC between results obtained on the low-level and high-level.

\begin{definition}[IC error wrt an intervention \cite{rischel2020category}] \label{def:ICerror} Given an abstraction $\absi$ from $\scm$ to $\scmi$, and given two disjoint sets $\mathbf{X'},\mathbf{Y'} \in \mathcal{X'}$, we define the IC error wrt the interventional distribution $P(\mathbf{Y'} \vert do(\mathbf{X'}))$ by considering the following diagram:

\begin{center}
\begin{tikzpicture}[shorten >=1pt, auto, node distance=1cm, thick, scale=1.0, every node/.style={scale=1.0}]

\node[] (M0_0) at (0,0) {$\scm[\mathrm{a}^{-1}(\mathbf{X'})]$};
\node[] (M0_1) at (4,0) {$\scm[\mathrm{a}^{-1}(\mathbf{Y'})]$};
\node[] (M1_0) at (0,-2) {$\scmi[\mathbf{X'}]$};
\node[] (M1_1) at (4,-2) {$\scmi[\mathbf{Y'}]$};

\draw[->]  (M0_0) to node[above,font=\small]{$\mu_{do(a^{-1}(\mathbf{X'}))}$} (M0_1);
\draw[->]  (M1_0) to node[above,font=\small]{$\nu_{do(\mathbf{X'})}$} (M1_1);
\draw[->]  (M0_0) to node[left,font=\small]{$\alpha_{\mathbf{X'}}$} (M1_0);
\draw[->]  (M0_1) to node[left,font=\small]{$\alpha_{\mathbf{Y'}}$} (M1_1);

\end{tikzpicture}
\end{center}

where $\mu_{do()}$, $\nu_{do()}$ are the stochastic functions computed in the respective interventional models, and by evaluating:
\begin{equation}\label{eq:EIC}
 \EIC{\absi,\mathbf{X'},\mathbf{Y'}} = D_{JSD}(\alpha_{\mathbf{Y'}} \circ \mu_{do(a^{-1}(\mathbf{X'}))}, \nu_{do(\mathbf{X'})} \circ \alpha_{\mathbf{X'}}),   
\end{equation}
where $D_{JSD}$ is the Jensen-Shannon distance (JSD) and $\circ$ denotes function composition.
\end{definition}

The definition of JSD is recalled in App. \ref{app:JSD}. This diagram evaluates the discrepancy between performing first an intervention on the low level and then abstracting, or abstracting first and then performing an intervention on the high level. Beyond a causal reading, the diagram has an algebraic and categorical reading. Algebraically, every node is associated with the set of outcomes of the given variable(s), while the stochastic functions and abstractions on the arrows can be expressed as matrices; this means that arrow composition can be efficiently computed by matrix multiplication; see App. \ref{app:algebraicdefinition} for further details. Categorically, the diagram has a rigorous meaning in the category $\mathtt{FinStoch}$ enriched in $\mathtt{Met}$ \cite{rischel2020category,fritz2020synthetic}.

We can extend the notion of IC error from an intervention to the abstraction itself:

\begin{definition}[Overall IC error \cite{rischel2020category}] \label{def:overallICerror}
Given an abstraction $\absi$ from $\scm$ to $\scmi$, the overall IC error is:
\begin{equation}
\eIC{\absi} = \sup_{(\mathbf{X'},\mathbf{Y'}) \in \mathcal{J}} \EIC{\absi,\mathbf{X'},\mathbf{Y'}},
\end{equation}
where $\mathcal{J}$ is the set of all non-empty disjoint pairs $(\mathbf{X'},\mathbf{Y'}) \in \mathscr{P}(\mathcal{X'})\times\mathscr{P}(\mathcal{X'})$, with $\mathscr{P}()$ being the powerset. 
\end{definition}

\begin{example} \label{ex:SCM0SCM1_abs}
Given the abstraction in Ex. \ref{ex:SCM0SCM1}, Lab A can measure the overall IC error $\eIC{\absi}\approx0.385$ wrt the set of pairs $\mathcal{J}=\{ (Sm',Hea') \}$. See App. \ref{app:samplemodels} for the exact computation.
\end{example}

\section{Measures of Abstraction Approximation}\label{sec:measures}

While IC provides a measure of interventional alignment between the low- and high-level model, this measure may not properly capture the priorities of an agent and weight candidate abstractions accordingly.

\begin{example}
Suppose Lab A with its model in Fig. \ref{fig:SCM0} is looking for an abstraction. Since abstraction $\absi$ to the model of Fig. \ref{fig:SCM1} has IC error $\eIC{\absi}\approx0.385$ as in Ex. \ref{ex:SCM0SCM1_abs}, Lab A may consider an abstraction $\absii$ to the singleton model in Fig. \ref{fig:SCMstar}. By definition, $\eIC{\absii}=0$. However, despite the lower error, this abstraction may be problematic given that the singleton model trivially carries no information.
\end{example}

To enrich our understanding of abstraction approximation, we use the definition of IC as a template for a generalizing the notion of \emph{error wrt an intervention} and \emph{overall error}.

\begin{definition}[Error wrt an intervention] \label{def:error} Given an abstraction $\absi$ from $\scm$ to $\scmi$, and given two disjoint sets $\mathbf{X'},\mathbf{Y'} \in \mathcal{X'}$, we define the error wrt the interventional distribution $P(\mathbf{Y'} \vert do(\mathbf{X'}))$ by considering the following diagram:

\begin{center}
\begin{tikzpicture}[shorten >=1pt, auto, node distance=1cm, thick, scale=1.0, every node/.style={scale=1.0}]

\node[] (M0_0) at (0,0) {$\scm[\mathrm{a}^{-1}(\mathbf{X'})]$};
\node[] (M0_1) at (4,0) {$\scm[\mathrm{a}^{-1}(\mathbf{Y'})]$};
\node[] (M1_0) at (0,-2) {$\scmi[\mathbf{X'}]$};
\node[] (M1_1) at (4,-2) {$\scmi[\mathbf{Y'}]$};

\draw[->]  (M0_0) to node[above,font=\small]{$\mu_{do(a^{-1}(\mathbf{X'}))}$} (M0_1);
\draw[->]  (M1_0) to node[above,font=\small]{$\nu_{do(\mathbf{X'})}$} (M1_1);
\draw[->,bend right]  (M0_0) to node[left,font=\small]{$\alpha_{\mathbf{X'}}$} (M1_0);
\draw[->,bend right]  (M0_1) to node[left,font=\small]{$\alpha_{\mathbf{Y'}}$} (M1_1);
\draw[->,bend right]  (M1_0) to node[right,font=\small]{$\mpinv{\alpha_{\mathbf{X'}}}$} (M0_0);
\draw[->,bend right]  (M1_1) to node[right,font=\small]{$\mpinv{\alpha_{\mathbf{Y'}}}$} (M0_1);
\end{tikzpicture}
\end{center}

where $\mpinv{\alpha_{\mathbf{X}'}}$ is the pseudo-inverse of $\alpha_{\mathbf{X}'}$, and by evaluating:
\begin{equation}\label{eq:EI}
 \EI{\absi,\mathbf{X'},\mathbf{Y'}} = D(p,q),   
\end{equation}
where $D$ is a distance, and $p,q$ are two paths in the above diagram with the same start and end points (as in Tab. \ref{tab:fourmeasures}).
\end{definition}


\begin{definition}[Overall error] \label{def:overallerror}
Given an abstraction $\absi$ from $\scm$ to $\scmi$, the overall interventional error is:
\begin{equation}
\eI{\absi} = \underset{(\mathbf{X'},\mathbf{Y'}) \in \mathcal{J}}{f} \EI{\absi,\mathbf{X'},\mathbf{Y'}},
\end{equation}
where $f$ is an aggregation function $f:\mathbb{R}^{|\mathcal{J}|} \rightarrow \mathbb{R}$, and $\mathcal{J}$ is an assessment set containing pairs $(\mathbf{X'},\mathbf{Y'})$.
\end{definition}

The definitions are generic and depend on five parameters:
\begin{itemize}
    \item $D$, the distance measure to assess an individual error $\EI{\absi,\mathbf{X'},\mathbf{Y'}}$. We rely on JSD which guarantees abstraction compositionality for IC \cite{rischel2020category} and for other interventional measures (see Sec. \ref{sec:properties}). Alternative measures guaranteeing the same property, such as $p$-Wasserstein distances, had already been suggested in \cite{rischel2020category,rischel2021compositional} and could be used in place of JSD.
    
    \item $\mpinv{\alpha_{\mathbf{X}'}}$, the pseudo-inverse of $\alpha_{\mathbf{X}'}$. We will adopt the standard Moore-Penrose inverse, whose definition and relevant properties are discussed in App. \ref{app:MPInverse}.
    
    \item $p,q$, the paths to be considered on the diagram in Definition \ref{def:error}; possible choices are discussed in Sec. \ref{ssec:Paths}.
    
    \item $f$, the function aggregating the individual errors $\EI{\absi,\mathbf{X'},\mathbf{Y'}}$. We adopt a supremum aggregation function, which provides a robust, worst-case scenario, evaluation of the error. Other functions may be considered in different scenarios, such as mean or weighted average. Moreover, ensembling theory \cite{dietterich2000ensemble,rokach2019ensemble} may help definining desirable properties for aggregation and analyzing correlations between errors.
    
    \item $\mathcal{J}$, the assessment set to evaluate $\eI{\absi}$; we will discuss possible choices in detail in Sec. \ref{ssec:Assessmentset}.
\end{itemize}

We focus on paths and assessment sets as they allow for the definition of new measures and crucially contribute to the meaning and the computational complexity of our measures.

\subsection{Paths} \label{ssec:Paths}
In the diagram of Definition \ref{def:error} the two horizontal arrows, $\mu_{do()}$ and $\nu_{do()}$, have a defined directionality; they capture causal mechanisms, and their inverse would represent anti-causal relationships, which are of no interest in this context. The two vertical arrows, however, may be considered in both directions: it may be desirable to move between a low-level and a high-level model in both ways. This naturally leads to the definition of four different measures, listed in Tab. \ref{tab:fourmeasures}, each one having relevance in light of specific settings and downstream tasks an agent may face. We present these measures by defining the paths $p,q$ and illustrating their use on an example using a lung cancer model from \cite{guyon2008design}.

\begin{table}
\begin{centering}
\begin{tabular}{cccc}
\hline 
 & {\color{blue}p} & {\color{red}q} & diagram \tabularnewline
\hline 
IC & $\nu_{do()} \circ \alpha_{\mathbf{X'}}$ & $ \alpha_{\mathbf{Y'}} \circ \mu_{do()}$ &
\begin{tikzpicture}[]
			\node[circle,fill,inner sep=1pt] (a) at (0,0) {};
			\node[circle,fill,inner sep=1pt] (b) at (0.5,0) {};
               \node[circle,fill,inner sep=1pt] (c) at (0,-0.5) {};
               \node[circle,fill,inner sep=1pt] (d) at (0.5,-0.5) {};
			
			\draw[->,red]  (a) to (b);
			\draw[->,blue]  (a) to (c);
                \draw[->,red]  (b) to (d);
			\draw[->,blue]  (c) to (d);
		\end{tikzpicture}
\tabularnewline
\hline 
IIL & $\mu_{do()}$ & $\mpinv{\alpha_{\mathbf{Y'}}} \circ \nu_{do()} \circ \alpha_{\mathbf{X'}}$ &
\begin{tikzpicture}[]
			\node[circle,fill,inner sep=1pt] (a) at (0,0) {};
			\node[circle,fill,inner sep=1pt] (b) at (0.5,0) {};
               \node[circle,fill,inner sep=1pt] (c) at (0,-0.5) {};
               \node[circle,fill,inner sep=1pt] (d) at (0.5,-0.5) {};
			
			\draw[->,blue]  (a) to (b);
			\draw[->,red]  (a) to (c);
                \draw[->,red]  (d) to (b);
			\draw[->,red]  (c) to (d);
		\end{tikzpicture} \tabularnewline
\hline 
ISIL & $\nu_{do()}$ & ${\alpha_{\mathbf{Y'}}} \circ \mu_{do()} \circ \mpinv{\alpha_{\mathbf{X'}}}$ &
\begin{tikzpicture}[]
			\node[circle,fill,inner sep=1pt] (a) at (0,0) {};
			\node[circle,fill,inner sep=1pt] (b) at (0.5,0) {};
               \node[circle,fill,inner sep=1pt] (c) at (0,-0.5) {};
               \node[circle,fill,inner sep=1pt] (d) at (0.5,-0.5) {};
			
			\draw[->,red]  (a) to (b);
			\draw[->,red]  (c) to (a);
                \draw[->,red]  (b) to (d);
			\draw[->,blue]  (c) to (d);
		\end{tikzpicture}\tabularnewline
\hline 
ISC & $\mpinv{\alpha_{\mathbf{Y'}}}\circ\nu_{do()}$ & $\mu_{do()} \circ  \mpinv{\alpha_{\mathbf{X'}}}$ &
\begin{tikzpicture}[]
			\node[circle,fill,inner sep=1pt] (a) at (0,0) {};
			\node[circle,fill,inner sep=1pt] (b) at (0.5,0) {};
               \node[circle,fill,inner sep=1pt] (c) at (0,-0.5) {};
               \node[circle,fill,inner sep=1pt] (d) at (0.5,-0.5) {};
			
			\draw[->,red]  (a) to (b);
			\draw[->,red]  (c) to (a);
                \draw[->,blue]  (d) to (b);
			\draw[->,blue]  (c) to (d);
		\end{tikzpicture}\tabularnewline
\hline 
\end{tabular}
\par\end{centering}
\caption{Interventional measures wrt different paths.}\label{tab:fourmeasures}
\end{table}

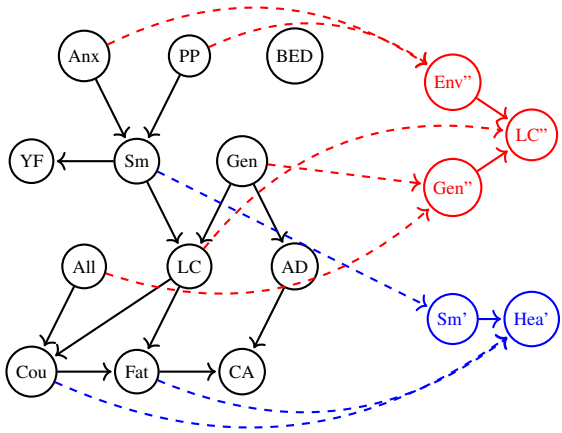
\begin{figure}
    \centering
		\begin{tikzpicture}[shorten >=1pt, auto, node distance=1cm, thick, scale=0.7, every node/.style={scale=0.7}]
			\tikzstyle{node_style} = [circle,draw=black]
			\node[node_style] (Anx) at (0,0) {Anx};
                \node[node_style] (PP) at (2,0) {PP};
			\node[node_style] (BED) at (4,0) {BED};

                \node[node_style] (YF) at (-1,-2) {YF};
                \node[node_style] (Smo) at (1,-2) {Sm};
			\node[node_style] (Gen) at (3,-2) {Gen};

                \node[node_style] (All) at (0,-4) {All};
                \node[node_style] (LC) at (2,-4) {LC};
			\node[node_style] (AD) at (4,-4) {AD};

                \node[node_style] (Cou) at (-1,-6) {Cou};
                \node[node_style] (Fat) at (1,-6) {Fat};
			\node[node_style] (CA) at (3,-6) {CA};

                \node[node_style,blue] (Smo_) at (7,-5) {Sm'};
                \node[node_style,blue] (Hea_) at (8.5,-5) {Hea'};

                \node[node_style,red] (Env_) at (7,-.5) {Env''};
		      \node[node_style,red] (Gen_) at (7,-2.5) {Gen''};
		      \node[node_style,red] (LC_) at (8.5,-1.5) {LC''};
   
			\draw[->]  (Anx) to (Smo);
                \draw[->]  (PP) to (Smo);
                \draw[->]  (Smo) to (YF);
                \draw[->]  (Smo) to (LC);
                \draw[->]  (Gen) to (AD);
                \draw[->]  (Gen) to (LC);
                \draw[->]  (LC) to (Cou);
                \draw[->]  (LC) to (Fat);
                \draw[->]  (All) to (Cou);
                \draw[->]  (Cou) to (Fat);
                \draw[->]  (AD) to (CA);
                \draw[->]  (Fat) to (CA);

                \draw[->,blue]  (Smo_) to (Hea_);

                \draw[->,blue,dashed]  (Smo) to (Smo_); 
                \draw[->,blue,dashed,bend right]  (Fat) to (Hea_);
                \draw[->,blue,dashed,bend right]  (Cou) to (Hea_);

                \draw[->,red]  (Env_) to (LC_);
    		\draw[->,red]  (Gen_) to (LC_);
    		
    		\draw[->,red,dashed,bend left]  (PP) to (Env_);
    		\draw[->,red,dashed,bend left]  (Anx) to (Env_);
    		\draw[->,red,dashed,bend left]  (LC) to (LC_);
    		\draw[->,red,dashed,bend right]  (All) to (Gen_);
    		\draw[->,red,dashed]  (Gen) to (Gen_);
                
		\end{tikzpicture}
		\caption{Base model (black), health model (blue) of Lab B from Fig. \ref{fig:SCM1} with the corresponding abstraction (dashed blue line), and lung cancer model (red) of Lab C from Fig. \ref{fig:SCM2} with the corresponding abstraction (dashed red line). Acronyms of variables are explained in Tab. \ref{tab:variablenames}.}
		\label{fig:examples}
\end{figure}

\subsubsection{Interventional Consistency (IC)} Discussed above, IC takes $p= \nu_{do()} \circ \alpha_{\mathbf{X'}}$ and $q = \alpha_{\mathbf{Y'}} \circ \mu_{do()}$. IC considers interventions on the low-level model and it evaluates the agreement, via abstraction, between results computed at the low-level and high-level. Low IC would be relevant when a downstream task depends on $P(\alpha_{\mathbf{Y'}}(\mathbf{Y}) \vert do(\mathbf{X}))$, with $\alpha_{\mathbf{Y'}}$ expressing a coarsening of $P(\mathbf{Y} \vert do(\mathbf{X}))$, while an agent wants to rely on the higher-level distribution $P(\mathbf{Y'} \vert do(\alpha_{\mathbf{X'}}(\mathbf{X}))$.

\begin{example} \label{ex:health1}
Consider the health scenario in Fig. \ref{fig:examples}, where Lab A has developed a large lung cancer SCM (black) and Lab B has produced a simpler model (blue). Lab A is performing a smoking experiment, estimating a health index from variables in its model, and feeding the result to a decision-making module to discriminate whether further exams are necessary.
Lab B wants to evaluate patients in its own model and provide results to the same decision-making module, concerned for patients to be processed equivalently in both models.
The original downstream task depends on $P(\alpha_{\mathbf{Y'}}(\mathbf{Y}) \vert do(\mathbf{X}))$, where $\mathbf{Y}$ is a set of variables aggregated in a health index via $\alpha_{\mathbf{Y'}}$ and $\mathbf{X}$ is smoking; Lab B will then be concerned with evaluating IC, thus reducing the discrepancy between $P(\alpha_{\mathbf{Y'}}(\mathbf{Y}) \vert do(\mathbf{X}))$ and $P(\mathbf{Y'} \vert do(\alpha_{\mathbf{X'}}(\mathbf{X}))$. The lower the IC, the more accurate the results (if the downstream decision-making module had been tuned on the low-level model) and the fairer the output (if we are concerned with the same proportions of patients being forwarded to further analysis by Lab A and Lab B).
\end{example}

It is worth remarking that, in this context we deal with interventional fairness, and not counterfactual fairness \cite{kusner2017counterfactual}; that is, fairness holds on the distributional level (models at different LAs produce the same interventional distributions), not the individual level (outcomes for an individual are not necessarily identical on different LAs). 

\subsubsection{Interventional Information Loss (IIL)} IIL takes $p= \mu_{do()}$ and $q = \mpinv{\alpha_{\mathbf{Y'}}} \circ \nu_{do()} \circ \alpha_{\mathbf{X'}}$. IIL also considers interventions on the low-level model and it evaluates the information lost by working through the high-level model. Low IIL would be relevant when a downstream task depends on $P(\mathbf{Y} \vert do(\mathbf{X}))$ but, because of constraints, an agent can not compute this quantity directly on the low-level model but it has to rely on coarser estimations obtained through the high-level model, thus ending to use $P(\mpinv{\alpha_{\mathbf{Y'}}}(\mathbf{Y'}) \vert do(\alpha_{\mathbf{X'}}(\mathbf{X}))$. In algebraic terms, IIL evaluates how well a low-level mechanisms may be decomposed into two abstractions and a high-level mechanism.

\begin{example} \label{ex:health2}
Consider the same health scenario as in Ex. \ref{ex:health1}. Lab A is performing a smoking experiment, estimating a health index, and predicting the probability of car accidents.
The downstream task can be described as depending on $P(\mathbf{Y} \vert do(\mathbf{X}))$, with $\mathbf{Y}$ being a set of variables for a health index and $\mathbf{X}$ smoking. Since estimating $\mathbf{Y}$ in its model is (in the relative terms of the example) computationally expensive, Lab A decides to rely on the model of Lab B. As the result will be re-used by Lab A for further computations, Lab A wants to estimate IIL, thus assessing the discrepancy between the expensive-to-compute $P(\mathbf{Y} \vert do(\mathbf{X}))$ and the cheaper $P(\mpinv{\alpha_{\mathbf{Y'}}}(\mathbf{Y'}) \vert do(\alpha_{\mathbf{X'}}(\mathbf{X}))$. This will guarantee that replacing some computations in the low-level model with higher-level model computations will produce results analogous to performing the whole computation at low-level.
\end{example}

Like fairness, replaceability between the original base model and the base model with a sub-part replaced holds only in an interventional, not counterfactual, sense.



\subsubsection{Interventional Superresolution Information Loss (ISIL)} ISIL takes $p= \nu_{do()}$ and $q = {\alpha_{\mathbf{Y'}}} \circ \mu_{do()} \circ \mpinv{\alpha_{\mathbf{X'}}}$. ISIL considers interventions on the high-level and it evaluates the information mismatch by working on the low-level model. Low ISIL would be relevant when a downstream task depends on $P(\mathbf{Y'} \vert do(\mathbf{X'}))$ but, because of constraints, an agent is requested to compute this quantity with higher precision on the low-level model, thus ending to rely on $P(\alpha_{\mathbf{Y'}}(\mathbf{Y}) \vert do(\mpinv{\alpha_{\mathbf{X'}}}(\mathbf{X'}))$. In a way complementary to IIL, ISIL evaluates how well a high-level mechanisms may be factored into two abstractions and a low-level mechanism.

\begin{example} \label{ex:lc1}
Consider the lung cancer scenario in Fig. \ref{fig:examples}, where Lab A has developed a large lung cancer SCM (black) and now Lab C has produced a simpler model (red). Lab C is performing environmental manipulation, and using the result to evaluate other high-level statistics that depend on $P(\mathbf{Y'} \vert do(\mathbf{X'}))$, with $\mathbf{Y'}$ being lung cancer and $\mathbf{X'}$ environment. 
Given the sensitivity of the evaluation, Lab C wants to match the more detailed model of Lab A by optimizing for ISIL, thus minimizing the discrepancy between its approximate $P(\mathbf{Y'} \vert do(\mathbf{X'}))$ and the finer-grained $P(\alpha_{\mathbf{Y'}}(\mathbf{Y}) \vert do(\mpinv{\alpha_{\mathbf{X'}}}(\mathbf{X'}))$. 
\end{example}

Notice that, because of the properties of the Moore-Penrose inverse, for an intervention $do(\mathbf{X'}=\mathbf{x'})$, $\mpinv{\alpha_{\mathbf{X'}}}$ entails a uniform distribution of probability mass over all $do(\mathbf{X}=\mathbf{x}) $ such that $a(\mathbf{X}) = \mathbf{X'}$ and $\alpha_{\mathbf{X'}}(\mathbf{x}) = \mathbf{x'}$. This uniform solution may be physically meaningless, and in \cite{rischel2021compositional} an abstraction is indeed enriched with an additional explicit map between high-level interventions and low-level interventions. 

\subsubsection{Interventional Superresolution Consistency (ISC)} ISC takes $p= \mpinv{\alpha_{\mathbf{Y'}}}\circ\nu_{do()}$ and $q = \mu_{do()} \circ  \mpinv{\alpha_{\mathbf{X'}}}$. ISC considers interventions on the high-level and it evaluates the agreement, via abstraction, between results computed at high-level and low-level. Lower ISC would be relevant when a downstream task depends on $P(\mpinv{\alpha_{\mathbf{Y'}}}(\mathbf{Y'}) \vert do(\mathbf{X'}))$, with $\mpinv{\alpha_{\mathbf{Y'}}}$ expressing a refinement of $P(\mathbf{Y'} \vert do(\mathbf{X'}))$, and an agent is required to work with $P(\mathbf{Y} \vert do(\mpinv{\alpha_{\mathbf{X'}}}(\mathbf{X'}))$.

\begin{example} \label{ex:lc2}
Consider the same lung cancer scenario as in Ex. \ref{ex:lc1}. Lab C is performing environmental manipulation, estimating the probability of lung cancer, and using a decision-making module to recommend further treatment.
As patients are undergoing a similar experiment in the more sophisticated model of Lab A, it is required that outcomes between the two models are aligned.
The downstream task depends on $P(\mpinv{\alpha_{\mathbf{Y'}}}(\mathbf{Y'}) \vert do(\mathbf{X'}))$, where $\mathbf{Y'}$ is lung cancer refined in the low-level model via $\mpinv{\alpha_{\mathbf{Y'}}}$ and $\mathbf{X'}$ is environment. Lab C will then evaluate ISC, which allows it to measure the discrepancy between $P(\mpinv{\alpha_{\mathbf{Y'}}}(\mathbf{Y'}) \vert do(\mathbf{X'}))$ and $P(\mathbf{Y} \vert do(\mpinv{\alpha_{\mathbf{X'}}}(\mathbf{X'}))$.
\end{example}

Tab. \ref{tab:pathtasks} summarizes the four interventional measures. 



\begin{table}
\begin{centering}
\begin{tabular}{ccc}
\hline 
& Original task & Abstraction task \tabularnewline
\hline 
\hline 
IC 
& $P({\alpha_{\mathbf{Y'}}}(\mathbf{Y}) \vert do(\mathbf{X}))$ & $P(\mathbf{Y'} \vert do(\alpha_{\mathbf{X'}}(\mathbf{X}))$ \tabularnewline
\hline 
IIL 
& $P(\mathbf{Y} \vert do(\mathbf{X}))$ & $P(\mpinv{\alpha_{\mathbf{Y'}}}(\mathbf{Y'}) \vert do(\alpha_{\mathbf{X'}}(\mathbf{X}))$ \tabularnewline
\hline 
ISIL 
& $P(\mathbf{Y'} \vert do(\mathbf{X'}))$ & $P(\alpha_{\mathbf{Y'}}(\mathbf{Y}) \vert do(\mpinv{\alpha_{\mathbf{X'}}}(\mathbf{X'}))$ \tabularnewline
\hline 
ISC 
& $P(\mpinv{\alpha_{\mathbf{Y'}}}(\mathbf{Y'}) \vert do(\mathbf{X'}))$ & $P(\mathbf{Y} \vert do(\mpinv{\alpha_{\mathbf{X'}}}(\mathbf{X'}))$ \tabularnewline
\hline
\end{tabular}
\par\end{centering}
\caption{Relation between interventional measures of abstraction approximation and downstream tasks. 
\emph{Original task} specifies on which distribution an original downstream task depends; \emph{Abstraction task} denotes on which distribution of a higher LA the task may depend.}\label{tab:pathtasks}

\end{table}

\subsection{Assessment Set}\label{ssec:Assessmentset}

The definition of an assessment set is crucial in the computation of an interventional measure of abstraction approximation. We discuss a few representative options, highlighting again their differences wrt potential downstream tasks. 


The choice to consider non-empty disjoint pairs of sets prevents us from considering observational distributions such as $P(\mathbf{X'})=P(\mathbf{X'}\vert do(\emptyset))$ corresponding to the pair $(\emptyset,\mathbf{X'})$. This increases the robustness of the measure to differences in the marginal distributions of the root-nodes, providing a degree of insensitivity to root covariate shift.

\begin{example}\label{ex:assessment1}
   Let Lab A and Lab B work with two SCMs having the same DAG as in Fig. \ref{fig:SCM1}, and let us assume an identity abstraction between them. As long as they specify the same mechanism $Sm' \rightarrow Hea'$, then $\eIC{\absi}=0$ independently from the marginal distributions on $Sm'$.
\end{example}

However, robustness to differences in observational marginal or to anti-causal quantities is only partial.

\begin{example}
Let us take the same setup as in Ex. \ref{ex:assessment1}, and assume Lab B wants to consider the error wrt the disjoint pair $(Hea',Sm')$. This would correspond to evaluating error wrt the interventional quantity $P(Sm' \vert do(Hea'))$. This presents two problems: (i) the stochastic matrix capturing this distribution would have an anti-causal meaning; and (ii) because of the form of the DAG, $P(Sm' \vert do(Hea'))=P(Sm')$, leading us to account for an observational quantity.
\end{example}

To avoid the error being affected by anti-causal quantities, we can define a causal assessment set:

\begin{definition}[Causal Assessment Set] \label{def:causalAssessmentSet}
Given an abstraction $\absi$ from $\scm$ to $\scmi$, let $\mathcal{J}_c$ be the set of all non-empty disjoint pairs $(\mathbf{X'},\mathbf{Y'}) \in \mathscr{P}(\mathcal{X'})\times\mathscr{P}(\mathcal{X'})$, such that $\forall Y' \in \mathbf{Y'}$, $\exists {X'} \prec {Y'}$ in $\scmi_{do(\mathbf{X'})}$. 
\end{definition}

The ordering relation guarantees that every node in the outcome $\mathbf{Y'}$ is affected by $\mathbf{X'}$. An equivalent definition in terms of independence is offered in App. \ref{app:independencebaseddef}.

Moreover, the evaluation is not, in general, robust to root covariate shift, as contributions from marginal distributions may always enter the evaluation if paths from the root nodes are not blocked.

\begin{example}
    Let Lab C work with an abstracted model as in Fig. \ref{fig:SCM2}, and assume it wants to consider the error wrt the disjoint pair $(Env'',LC'')$. This would correspond to evaluating error wrt the interventional quantity $P(LC'' \vert do(Env''))$. In the model, this corresponds to $\sum_{Gen''} P(LC'' \vert Env'',Gen'')P(Gen'')$, revealing the contribution of the marginal $P(Gen'')$.
\end{example}

This sensitivity may be desirable in certain cases, for instance when the contribution of marginals can not be suppressed or when deemed informative. However, if we were interested in assessing abstraction error robustly wrt covariate shift, that is, only wrt the actual causal mechanisms, we could consider a parental assessment set.

\begin{definition}[Parental Assessment Set] \label{def:parentalAssessmentSet}
Given an abstraction $\absi$ from $\scm$ to $\scmi$, let $\mathcal{J}_p$ be the set of all non-empty disjoint pairs $(\mathbf{X'},\mathbf{Y'}) \in \mathscr{P}(\mathcal{X'})\times\mathscr{P}(\mathcal{X'})$, such that $\mathbf{X'} = Pa(\mathbf{Y'})$.
\end{definition}

Lastly, it may be worth pointing out that custom assessment sets $\mathcal{J}_u$ may always be defined by an agent, if it is interested in the abstraction only of specific sub-parts of a base model. Tab. \ref{tab:assessmenttasks} summarizes the assessment sets we considered, and the type of relevant downstream tasks of interest.

\begin{table}
\begin{centering}
\begin{tabular}{c>{\centering}p{5cm}}
{Assessment set} & {Downstream task dependes on...}\tabularnewline
\hline
\hline
Complete $\mathcal{J}$ & any possible causal or anti-causal intervention $P(\mathbf{Y'}\vert do(\mathbf{X'}))$\tabularnewline
\hline
Causal $\mathcal{J}_{c}$ & any possible causal intervention $P(\mathbf{Y'}\vert do(\mathbf{X'}))$,
potentially affected by root covariate shift\tabularnewline
\hline
Parental $\mathcal{J}_{p}$ & causal intervention $P(\mathbf{Y'}\vert do(\mathbf{X'}))$ dependent only on causal mechanisms\tabularnewline
\hline
Custom $\mathcal{J}_{u}$ & user-specific set of interventions $P(\mathbf{Y'}\vert do(\mathbf{X'}))$ \tabularnewline
\hline
\end{tabular}
\par\end{centering}
\begin{centering}
\caption{Relation between assessment sets and downstream tasks.}\label{tab:assessmenttasks}
\par\end{centering}
\end{table}

\section{Properties of Abstraction Approximation} \label{sec:properties}
All interventional measures of abstraction approximation share common properties, relevant for learning abstractions. Complete proofs are provided in App. \ref{app:errorproperties} and \ref{app:overallerrorproperties}.

\subsection{Properties of Error}

Let us consider two generic abstractions: $\absi$ from $\scm$ to $\scmi$, and $\absii$ from $\scmi$ to $\scmii$. 
Thanks to JSD, a key property is (E1) \emph{triangle inequality}, that is, $\EI{\absii\absi,\mathbf{X''},\mathbf{Y''}} \leq \EI{\absi,\mathbf{X'},\mathbf{Y'}} + \EI{\absii,\mathbf{X''},\mathbf{Y''}}$, which is grounded in horizontal compositionality (or composition of abstractions) \cite{rischel2020category}.
Two other forms of compositionality instead do not hold: (NE1) \emph{vertical non-compositionality} (or stochastic function non-composition); and (NE2) \emph{product non-compositionality}. These negative properties follow from stochastic functions being computed from different post-interventional models (see App. \ref{app:errorproperties} for a more precise discussion).
From (E1) we immediately derive (E2) \emph{non-monotonicity}, stating that it is not guaranteed that $\EI{\absii\absi,\mathbf{X''},\mathbf{Y''}} \geq \EI{\absi,\mathbf{X'},\mathbf{Y'}}$. From the definition we can identify extrema of the error: (E3) \emph{zero at identity}, when the abstracted model is an identity; (E4) \emph{zero at singleton}, only for IC, when the abstracted model is a singleton.


Finally, we have properties concerning relationships and identities (see App. \ref{app:errorproperties}) between interventional measures.

\begin{proposition}[Relationship between measures] \label{prop:relationships}
We have a partial ordering among the interventional measures as:
$\EIIL{\absi,\mathbf{X'},\mathbf{Y'}} \geq \EIC{\absi,\mathbf{X'},\mathbf{Y'}}$, 
$\EIIL{\absi,\mathbf{X'},\mathbf{Y'}} \geq \EISC{\absi,\mathbf{X'},\mathbf{Y'}}$,
$\EIC{\absi,\mathbf{X'},\mathbf{Y'}} \geq \EISIL{\absi,\mathbf{X'},\mathbf{Y'}}$,
$\EISC{\absi,\mathbf{X'},\mathbf{Y'}} \geq \EISIL{\absi,\mathbf{X'},\mathbf{Y'}}$.
\end{proposition}

\emph{Proof sketch.} All relations can be proved by applying the property of shortness of JSD and the right inverse property of the Moore-Penrose inverse.  $\blacksquare$


\subsection{Properties of Overall Error}
Properties of the error may immediately extend to the overall error according to the chosen aggregation function $f$. In the case of the supremum, the overall error inherits the properties of: (O1) \emph{triangle inequality}; (O2) \emph{non-monotonicity}; (O3) \emph{zero at identity}; and (O4) \emph{zero at singleton}.
Also, extension of Proposition \ref{prop:relationships} hold; however, notice that, despite this extension, two interventional measures of abstraction approximation may reach their minima for different abstractions. For instance, given two abstractions $\absi$ and $\absii$ from $\scm$ to $\scmi$, such that $\eIIL{\absi} \leq \eIIL{\absii}$, while it holds that $\eIIL{\absi} \geq \eIC{\absi}$ and $\eIIL{\absii} \geq \eIC{\absii}$, it does not follow that $\eIC{\absi} \leq \eIC{\absii}$; so different abstractions between the same two SCMs $\scm$ and $\scmi$ may minimize different interventional measures. 

A necessary condition for any error measure to be finite is:

\begin{proposition}[Finiteness of the overall error] \label{prop:finiteness}
$\eI{\absi} < \infty$ if $a$ is order-preserving.
\end{proposition}

\emph{Proof sketch.} It can be shown that, in absence of order-preservation, it is impossible to compose the paths required for computing any error measure.  $\blacksquare$
 

This proposition holds for all measures because it is related to the directionality of the edges representing causal mechanisms. This condition implicitly asserts that an abstraction can not reverse the directionality of causation, a requirement explicit in certain abstraction frameworks \cite{otsuka2022equivalence}.
Also, the request of order-preservation has a connection to the framework of \cite{rubenstein2017causal}, where order-preservation is imposed on a map $\omega$ relating low-level interventions with high-level interventions. 
Imposing order-preservation between variables acts at a more basic level and implies order-preservation among the interventions.

\section{Abstraction Evaluation and Learning}\label{sec:evaluating}

A simple algorithm for evaluating abstraction can be derived from the original specification of IC error in Definition \ref{def:overallICerror} by computing the error for all pairs $(\mathbf{X'},\mathbf{Y'}) \in \mathcal{J}$, as in Alg. \ref{alg:eICevaluator}. Its complexity depends on the loop in step 3: at each iteration, two matrix multiplications and a JSD computation are performed in order to evaluate $\EIC{\absi,\mathbf{X},\mathbf{Y}}$. The overall complexity then grows as $O(|\mathcal{J}|)$, which, in the base case is given by the product of two powersets,  $|\mathcal{J}| \approx 2^{2|\mathcal{X'}|}$. 
Computational complexity can then be reduced by exploiting the structure of a SCM and shrinking the set $\mathcal{J}$. A causal assessment set relies on partial ordering to reduce the size $|\mathcal{J}_c| < |\mathcal{J}|$. Even more, a parental assessment set exploits parental relationships to set the size $|\mathcal{J}_p| = |\mathcal{X'}| < |\mathcal{J}_c|$. Custom assessment sets $\mathcal{J}_u$ may also limit the number of pairs to be considered.
Moreover, Prop. \ref{prop:finiteness} provides an efficient test to evaluate whether any interventional measure of abstraction approximation will be finite with complexity $O(|E|)$ proportional to the number of edges in the DAG $\mathcal{G}_{\scm}$; if not, no further computation is required. We can then evaluate the overall error more efficiently as in Alg. \ref{alg:eIevaluator}.
The main challenge in further shrinking the assessment set $\mathcal{J}$ follows from the negative properties (NE1, NE2): without compositionality, it is not possible to reduce the evaluation of composed interventions to the one of its composing parts. 

\begin{algorithm}[tb]
    \caption{Overall IC error evaluation}
    \label{alg:eICevaluator}
    \textbf{In}: $\scm$, $\scmi$, $\absi = \langle R,a,\alpha \rangle$\\
    \textbf{Out}: $\eIC{\absi}$
    \begin{algorithmic}[1] 
        \STATE Initialize $\mathbf{E}=\{\}$
        \STATE Let $\mathcal{J}$ be the set of all non-empty disjoint pairs $(\mathbf{X'},\mathbf{Y'})$
        \FOR[$O(2^{2|\mathcal{X'}|})$]{$(\mathbf{X'},\mathbf{Y'}) \in \mathcal{J}$}
            \STATE Compute $\EIC{\absi,\mathbf{X'},\mathbf{Y'}}$ as in Eq. \ref{eq:EIC} and add to $\mathbf{E}$
        \ENDFOR
        \STATE \textbf{return} $\sup \mathbf{E}$
    \end{algorithmic}
\end{algorithm}



\begin{algorithm}[tb]
    \caption{Abstraction evaluation}
    \label{alg:eIevaluator}
    \textbf{In}: $\scm$, $\scmi$, $\absi = \langle R,a,\alpha \rangle$, ${I} \in \{IC,IIL,ISIL,ISC\}$, $\mathcal{J}$\\
    \textbf{Out}: $\eI{\absi}$
    \begin{algorithmic}[1] 
        \IF[$O(|E|)$] {$a$ is not order-preserving}
            \STATE \textbf{return} $\infty$
        \ENDIF
        \STATE Initialize $\mathbf{E}=\{\}$
        

        \FOR[$O(|\mathcal{J}|)$]{$(\mathbf{X'},\mathbf{Y'}) \in \mathcal{J}$} 
            \STATE Compute $\EI{\absi,\mathbf{X'},\mathbf{Y'}}$ as in Eq. \ref{eq:EI} and add to $\mathbf{E}$
        \ENDFOR
        \STATE \textbf{return} $\sup \mathbf{E}$
    \end{algorithmic}
\end{algorithm}


Beyond evaluating abstractions, an agent may be interested in  improving or learning new abstractions.
If given an unsatisfactory abstraction $\absi$ with a high error $\eI{\absi}$, properties (O1) and (O2) point to the possibility of sequentially improving the abstraction by searching for a new abstraction $\absii$ such $\eI{\absii\absi} < \eI{\absi}$.
If given an incomplete abstraction $\absi$ for which one or more elements among $\scmi,R,a,\alpha_{X'}$ are not completely specified, it is possible to learn a complete specification of the abstraction by minimizing a chosen measure of abstraction approximation. In this case, properties (O3) and (O4) highlight minima the optimization may achieve.
In both cases, Alg. \ref{alg:eIevaluator} can be used as a building block to find an optimal abstraction by computing the error for each candidate abstraction in a set $\mathcal{K}$ (see Alg.  \ref{alg:eIlearner} in App. \ref{app:algorithms}). This algorithm has a computational complexity of $O(|\mathcal{K}||\mathcal{J}|)$.
 
\section{Empirical Evaluation}\label{sec:evaluation}

We run empirical simulations for the two scenarios in Fig. \ref{fig:examples}: (i) in the \emph{health scenario} we perform \emph{abstraction learning} and show how our metrics would produce different results fit for distinct downstream tasks; and, (ii) in the \emph{lung cancer scenario} we run \emph{abstraction evaluation} to show the effects of the choice of assessment sets. Base model details are provided in App. \ref{app:experimentaldetails}.
These scenarios are designed to encompass a variety of configurations such as 
different structures in the low- and high-level model (chains, colliders, and forks), different $a$-maps among nodes (one-to-one, many-to-one), different number of variables and domain cardinality in the high-level model.
Empirical distributions are computed from $10^4$ samples; means and standard deviations are computed out of $10$ repetitions.
All simulations are available online\footnote{\url{https://github.com/FMZennaro/CausalAbstraction/tree/main/papers/2023-quantifying-consistency-and-infoloss}}. 
 

\paragraph{Health scenario.}
Let's consider the health scenario in Fig. \ref{fig:examples}. The abstracted model (blue) is not fully defined (three candidate stochastic matrices for $Sm' \rightarrow Hea'$ have been proposed) and the abstraction itself is defined only in terms of $R$ and $a$. Lab B wants to learn the best stochastic matrix and abstraction, wrt interventions performed by Lab A, in light of the downstream tasks described in Ex. \ref{ex:health1} and \ref{ex:health2}. 
Lab B performs \emph{abstraction learning} using Alg. \ref{alg:eIlearner}. Two different solutions are learned by minimizing either IC or ILL. Definition of models, abstractions, and solutions are in App. \ref{app:heathscenario}.

Considering the downstream task described in Ex. \ref{ex:health1}, Tab. \ref{tab:health1} shows that 
the closest agreement between the low- and high-level model in classifying patients for further exam is achieved by minimizing IC.
Instead, considering the downstream task described in Ex. \ref{ex:health2}, Tab. \ref{tab:health2} 
shows that the best match between the predictive distribution computed in the low-level model and in the low-level model when replacing a sub-part of it with a high-level abstraction, is obtained by minimizing IIL. In conclusion, an agent should choose carefully which measure to minimize according to its aim.

\begin{table}
\begin{centering}
\begin{tabular}{ccc}
\hline 
 & {Opt IC} & {Opt IIL}\tabularnewline
\hline 
$\hat{P}(\mathbf{a}\vert do(Sm=0))$ & 0.607$\pm$0.003 & 0.413$\pm$0.006 \tabularnewline
$\hat{P}(\mathbf{b}\vert do(\alpha_{Sm'}(Sm)=0))$ & 0.600$\pm$0.003 & 0.498$\pm$0.006\tabularnewline
\hline 
$\hat{P}(\mathbf{a}\vert do(Sm=1))$ & 0.797$\pm$0.003  & 0.681$\pm$0.005  \tabularnewline
$\hat{P}(\mathbf{b}\vert do(\alpha_{Sm'}(Sm)=1))$ &   0.797$\pm$0.004 &   0.799$\pm$0.004\tabularnewline
\hline 
\end{tabular}
\par\end{centering}
\caption{Comparison between distributions $\hat{P}(\alpha_{\mathbf{Y'}}(\mathbf{Y}) \vert do(\mathbf{X}))$ and $\hat{P}(\mathbf{Y'} \vert do(\alpha_{\mathbf{X'}}(\mathbf{X})))$, where $\mathbf{Y'} = Hea'$ and $\mathbf{X'} = Sm'$, when learning by optimizing for IC or IIL. $\mathbf{a},\mathbf{b}$ are placeholders for $\alpha_{Hea'}(Hea)=1$ and $Hea'=1$, respectively.}\label{tab:health1}
\end{table}

\begin{table}
		\begin{centering}
			\begin{tabular}{ccc}
				\hline 
				& {$Sm=0$} & {$Sm=1$}\tabularnewline
				\hline 
				$\hat{P}(CA=1 \vert do(Sm))$ & 0.679$\pm$0.004 & 0.766$\pm$0.005 \tabularnewline
				$\hat{P}_{IC}(CA=1 \vert do(Sm))$ & 0.256$\pm$0.006 & 0.341$\pm$0.005\tabularnewline
				$\hat{P}_{IIL}(CA=1 \vert do(Sm))$ & 0.427$\pm$0.005  & 0.680$\pm$0.005  \tabularnewline
				\hline 
			\end{tabular}
			\par\end{centering}
	\caption{Comparison between the empirical distribution computed only on the low-level model $\hat{P}$ and distributions using the abstraction minimizing IC ($\hat{P}_{IC}$) or IIL ($\hat{P}_{IC}$). 
 }\label{tab:health2}
\end{table}

\paragraph{Lung cancer scenario.}
Let's consider the lung cancer scenario in Fig. \ref{fig:examples}. All models are completely defined, while the abstraction is given only in terms of $R$ and $a$. Lab C wants to find the best abstraction minimizing ISIL, in light of the downstream task described in Ex. \ref{ex:lc1} and while considering three different assessment sets (causal, parental and custom). Lab C performs \emph{abstraction evaluation} using Alg. \ref{alg:eIevaluator}. Three different solutions are learned by minimizing ISIL with the three assessment sets. Exact definition of models, abstractions and solutions are provided in App. \ref{app:lungcancerscenario}.

Tab. \ref{tab:lc1} confirms that,
if the aim is to predict lung cancer under environmental experiments, then the best result is obtained when minimizing wrt a targeted assessment sets ($\mathcal{J}_p, \mathcal{J}_u$); larger sets require more computation (see Tab. \ref{tab:lc1-time}), and end up selecting a solution that, by mediating among many interventions, underperforms wrt the intervention of interest. This demonstrates the importance for an agent to optimize wrt a set of interventions that is relevant to its aim.

\begin{table}
		\begin{centering}
			\begin{tabular}{cccc}
				\hline 
				& {$Env''=0$} & {$Env''=1$} & {$Env''=2$} \tabularnewline
				\hline
                    $\hat{P}(\mathbf{a})$ & 0.445$\pm$0.003 & 0.555$\pm$0.003 & 0.655$\pm$0.004 \tabularnewline
				$\hat{P}_{\mathcal{J}_{c}}(\mathbf{b})$ & 0.194$\pm$0.003 & 0.271$\pm$0.005 & 0.438$\pm$0.005 \tabularnewline
				$\hat{P}_{\mathcal{J}_{p}}(\mathbf{b})$ & 0.563$\pm$0.005 & 0.730$\pm$0.005 & 0.807$\pm$0.003 \tabularnewline
				$\hat{P}_{\mathcal{J}_{u}}(\mathbf{b})$ & 0.557$\pm$0.005  & 0.730$\pm$0.005 & 0.806$\pm$0.004   \tabularnewline
				\hline 
			\end{tabular}
			\par\end{centering}
	\caption{Comparison between the empirical distribution computed only on the high-level model ($\hat{P}$) and using the abstraction minimizing ISIL wrt causal set ($\hat{P}_{\mathcal{J}_c}$), parental set ($\hat{P}_{\mathcal{J}_p}$), or custom set ($\hat{P}_{\mathcal{J}_u}$). $\mathbf{a},\mathbf{b}$ are placeholders for $LC''=1 \vert do(Env'')$ and $\alpha_{LC''}(LC)=1 \vert do(\mpinv{\alpha_{Env''}}(Env''))$, respectively. 
 }\label{tab:lc1}
\end{table}



\section{Conclusion}\label{sec:conclusion}

We introduced a family of interventional measures of abstraction approximation to quantify consistency and information loss. 
Our empirical simulations show that optimizing such measures lead to learning different optimal abstractions, which fit different constraints and downstream tasks. The proposed framework empowers modellers and agents by providing them with a set of measures that will help them learn abstractions that better suit their specific needs.

While this work focuses on four key measures (IC, IIL, ISC, ISIL), the proposed framework can accommodate new custom measures by modifying one or more of the parameters discussed in Sec. \ref{sec:measures}, combinations of existing measures, or shifting the focus to observational/counterfactual properties.

Future work could consider enhancing our learning algorithm. In one direction, we want to exploit the structure of the SCMs and properties of graph morphisms to reduce the size of assessment sets $\mathcal{J}$. Formal bounds may also be found for the tradeoff between the size of $\mathcal{J}$ and the error in estimating $\EI{\absi}$. In another direction, we want to consider the algebraic properties of factoring stochastic matrices and sparse abstraction matrices in order to simplify the search space.





\section*{Acknowledgments}

TD acknowledges support from a UKRI Turing AI acceleration Fellowship [EP/V02678X/1]. The authors thank the anonymous reviewers for their suggestions in improving this work.

\bibliographystyle{named}
\bibliography{abstraction}

\begin{thebibliography}{}

\bibitem[\protect\citeauthoryear{Beckers and
  Halpern}{2019}]{beckers2018abstracting}
Sander Beckers and Joseph~Y Halpern.
\newblock Abstracting causal models.
\newblock In {\em Proceedings of the AAAI Conference on Artificial
  Intelligence}, volume~33, pages 2678--2685, 2019.

\bibitem[\protect\citeauthoryear{Beckers \bgroup \em et al.\egroup
  }{2020}]{beckers2020approximate}
Sander Beckers, Frederick Eberhardt, and Joseph~Y Halpern.
\newblock Approximate causal abstractions.
\newblock In {\em Uncertainty in Artificial Intelligence}, pages 606--615.
  PMLR, 2020.

\bibitem[\protect\citeauthoryear{Chalupka \bgroup \em et al.\egroup
  }{2017}]{chalupka2017causal}
Krzysztof Chalupka, Frederick Eberhardt, and Pietro Perona.
\newblock Causal feature learning: an overview.
\newblock {\em Behaviormetrika}, 44(1):137--164, 2017.

\bibitem[\protect\citeauthoryear{Chang and Fung}{1990}]{chang1990refinement}
Kuo-Chu Chang and Robert~M. Fung.
\newblock Refinement and coarsening of bayesian networks.
\newblock In {\em Conference on Uncertainty in Artificial Intelligence}, 1990.

\bibitem[\protect\citeauthoryear{Cover}{1999}]{cover1999elements}
Thomas~M Cover.
\newblock {\em Elements of information theory}.
\newblock John Wiley \& Sons, 1999.

\bibitem[\protect\citeauthoryear{Dietterich}{2000}]{dietterich2000ensemble}
Thomas~G Dietterich.
\newblock Ensemble methods in machine learning.
\newblock In {\em International workshop on multiple classifier systems}, pages
  1--15. Springer, 2000.

\bibitem[\protect\citeauthoryear{Eberhardt and Lee}{2022}]{eberhardt2022causal}
Frederick Eberhardt and Lin~Lin Lee.
\newblock Causal emergence: When distortions in a map obscure the territory.
\newblock {\em Philosophies}, 7(2):30, 2022.

\bibitem[\protect\citeauthoryear{Elidan and Gould}{2008}]{elidan2008learning}
Gal Elidan and Stephen Gould.
\newblock Learning bounded treewidth bayesian networks.
\newblock {\em Advances in neural information processing systems}, 21, 2008.

\bibitem[\protect\citeauthoryear{Fritz}{2020}]{fritz2020synthetic}
Tobias Fritz.
\newblock A synthetic approach to markov kernels, conditional independence and
  theorems on sufficient statistics.
\newblock {\em Advances in Mathematics}, 370:107239, 2020.

\bibitem[\protect\citeauthoryear{Geiger \bgroup \em et al.\egroup
  }{2021}]{geiger2021causal}
Atticus Geiger, Hanson Lu, Thomas Icard, and Christopher Potts.
\newblock Causal abstractions of neural networks.
\newblock {\em Advances in Neural Information Processing Systems},
  34:9574--9586, 2021.

\bibitem[\protect\citeauthoryear{Guyon \bgroup \em et al.\egroup
  }{2008}]{guyon2008design}
Isabelle Guyon, Constantin Aliferis, Greg Cooper, Andr{\'e} Elisseeff,
  Jean-Philippe Pellet, Peter Spirtes, and Alexander Statnikov.
\newblock Design and analysis of the causation and prediction challenge.
\newblock In {\em Causation and Prediction Challenge}, pages 1--33. PMLR, 2008.

\bibitem[\protect\citeauthoryear{Hoel}{2017}]{hoel2017map}
Erik~P Hoel.
\newblock When the map is better than the territory.
\newblock {\em Entropy}, 19(5):188, 2017.

\bibitem[\protect\citeauthoryear{Kroer and Sandholm}{2018}]{KroerS18}
Christian Kroer and Tuomas Sandholm.
\newblock A unified framework for extensive-form game abstraction with bounds.
\newblock In Samy Bengio, Hanna~M. Wallach, Hugo Larochelle, Kristen Grauman,
  Nicol{\`{o}} Cesa{-}Bianchi, and Roman Garnett, editors, {\em Advances in
  Neural Information Processing Systems 31: Annual Conference on Neural
  Information Processing Systems 2018, NeurIPS 2018, December 3-8, 2018,
  Montr{\'{e}}al, Canada}, pages 613--624, 2018.

\bibitem[\protect\citeauthoryear{Kusner \bgroup \em et al.\egroup
  }{2017}]{kusner2017counterfactual}
Matt~J Kusner, Joshua Loftus, Chris Russell, and Ricardo Silva.
\newblock Counterfactual fairness.
\newblock {\em Advances in neural information processing systems}, 30, 2017.

\bibitem[\protect\citeauthoryear{Levin}{1992}]{levin1992problem}
Simon~A Levin.
\newblock The problem of pattern and scale in ecology: the robert h. macarthur
  award lecture.
\newblock {\em Ecology}, 73(6):1943--1967, 1992.

\bibitem[\protect\citeauthoryear{Mitchell}{2021}]{mitchell2021abstraction}
Melanie Mitchell.
\newblock Abstraction and analogy-making in artificial intelligence.
\newblock {\em Annals of the New York Academy of Sciences}, 1505(1):79--101,
  2021.

\bibitem[\protect\citeauthoryear{Otsuka and
  Saigo}{2022}]{otsuka2022equivalence}
Jun Otsuka and Hayato Saigo.
\newblock On the equivalence of causal models: A category-theoretic approach.
\newblock {\em arXiv preprint arXiv:2201.06981}, 2022.

\bibitem[\protect\citeauthoryear{Pearl}{2009}]{pearl2009causality}
Judea Pearl.
\newblock {\em Causality}.
\newblock Cambridge University Press, 2009.

\bibitem[\protect\citeauthoryear{Penrose}{1955}]{penrose1955generalized}
Roger Penrose.
\newblock A generalized inverse for matrices.
\newblock In {\em Mathematical proceedings of the Cambridge philosophical
  society}, volume~51, pages 406--413. Cambridge University Press, 1955.

\bibitem[\protect\citeauthoryear{Rischel and
  Weichwald}{2021}]{rischel2021compositional}
Eigil~F Rischel and Sebastian Weichwald.
\newblock Compositional abstraction error and a category of causal models.
\newblock {\em arXiv preprint arXiv:2103.15758}, 2021.

\bibitem[\protect\citeauthoryear{Rischel}{2020}]{rischel2020category}
Eigil~Fjeldgren Rischel.
\newblock The category theory of causal models.
\newblock 2020.

\bibitem[\protect\citeauthoryear{Rokach}{2019}]{rokach2019ensemble}
Lior Rokach.
\newblock {\em Ensemble learning: pattern classification using ensemble
  methods}.
\newblock World Scientific, 2019.

\bibitem[\protect\citeauthoryear{Rubenstein \bgroup \em et al.\egroup
  }{2017}]{rubenstein2017causal}
Paul~K Rubenstein, Sebastian Weichwald, Stephan Bongers, Joris~M Mooij, Dominik
  Janzing, Moritz Grosse-Wentrup, and Bernhard Sch{\"o}lkopf.
\newblock Causal consistency of structural equation models.
\newblock In {\em 33rd Conference on Uncertainty in Artificial Intelligence
  (UAI 2017)}, pages 808--817. Curran Associates, Inc., 2017.

\bibitem[\protect\citeauthoryear{Sandholm}{2015}]{Sandholm15a}
Tuomas Sandholm.
\newblock Abstraction for solving large incomplete-information games.
\newblock In Blai Bonet and Sven Koenig, editors, {\em Proceedings of the
  Twenty-Ninth {AAAI} Conference on Artificial Intelligence, January 25-30,
  2015, Austin, Texas, {USA}}, pages 4127--4131. {AAAI} Press, 2015.

\bibitem[\protect\citeauthoryear{Schmutz \bgroup \em et al.\egroup
  }{2020}]{schmutz2020mesoscopic}
Valentin Schmutz, Wulfram Gerstner, and Tilo Schwalger.
\newblock Mesoscopic population equations for spiking neural networks with
  synaptic short-term plasticity.
\newblock {\em The Journal of Mathematical Neuroscience}, 10(1):1--32, 2020.

\bibitem[\protect\citeauthoryear{Spivak}{2014}]{spivak2014category}
David~I Spivak.
\newblock {\em Category theory for the sciences}.
\newblock MIT Press, 2014.

\bibitem[\protect\citeauthoryear{Zennaro \bgroup \em et al.\egroup
  }{2023}]{zennaro2023jointly}
Fabio~Massimo Zennaro, M{\'a}t{\'e} Dr{\'a}vucz, Geanina Apachitei, W.~Dhammika
  Widanage, and Theodoros Damoulas.
\newblock Jointly learning consistent causal abstractions over multiple
  interventional distributions.
\newblock In {\em 2nd Conference on Causal Learning and Reasoning}, 2023.

\end{thebibliography}

\newpage

\appendix

\twocolumn[{%
 \centering
 \LARGE \vspace{2.5em}\textbf{Appendices to\\ Quantifying Consistency and Information Loss for Causal Abstraction Learning}\vspace{3.5em}
}]

\section{JSD} \label{app:JSD}

\begin{definition*}[Jensen-Shannon distance (JSD)]
    Given two probability mass functions $p$ and $q$ on the same domain $\mathcal{X}$, such that $p(x)>0$ and $q(x)>0$ $\forall x\in\mathcal{X}$, the JSD is:
\begin{equation*}
    D_{JSD}(p,q) = \sqrt{\frac{1}{2} d_{KL}(p;m) + \frac{1}{2} d_{KL}(q;m)},
\end{equation*}
where $d_{KL}(p,q) = -\sum_{x\in\mathcal{X}} p(x) \log \frac{p(x)}{q(x)}$ is the Kullback-Leibler (KL) divergence,  and $m = \frac{p+q}{2}$.
\end{definition*}  

JSD has the following properties:
\begin{itemize}
    \item[(J1)] {Non-negativity:} $D_{JSD}(p,q) \geq 0$.
    \item[(J2)] {Identity of the indiscernibles:} $D_{JSD}(p,q) = 0$ iff $p=q$.
    \item[(J3)] {Symmetry:} $D_{JSD}(p,q) = D_{JSD}(q,p)$.
    \item[(J4)] {Triangle inequality:} $D_{JSD}(p,q) \leq D_{JSD}(p,r) + D_{JSD}(r,q)$.
    \item[(J5)] {Shortness:} $D_{JSD}(p,q) \geq D_{JSD}(fp,fq)$ and $D_{JSD}(p,q) \geq D_{JSD}(pf,qf)$ for any deterministic $f$.
\end{itemize}
Properties (J1)-(J4) are standard properties of any distance metric \cite{cover1999elements}. Property (J5) follows from the definition of JSD \cite{rischel2020category}; intuitively, it means that pre-processing or post-processing two distributions by the same deterministic function can only reduce their JSD.

\section{Algebraic Defintion of the Commuting Diagram} \label{app:algebraicdefinition}

A diagram as in Definition \ref{def:ICerror} can be given an algebraic encoding in the following way.
\begin{itemize}
    \item A nodes $\scm[\mathbf{X}]$ can be encoded as a set, specifically the set of outcomes associated with the endogenous variable(s) $\mathbf{X}$, as specified in the definition of the SCM $\scm$.

    \item A horizontal arrow $\mu_{do(\mathbf{X})}: \scm[\mathbf{X}] \rightarrow \scm[\mathbf{Y}]$ expresses a stochastic function between the set of outcomes of endogenous variable(s) $\mathbf{X}$ and endogenous variable(s) $\mathbf{Y}$. Since both domains are finite (by Definition \ref{def:SCM}), the stochastic function can be encoded in a (column-stochastic) matrix with dimension $|\scm[\mathbf{Y}]| \times |\scm[\mathbf{X}]|$. Intuitively, every column encodes a distribution over the outcomes $\scm[\mathbf{Y}]$ for a given value of an intervention on $\scm[\mathbf{X}]$. Notice that a stochastic function $\mu_{do(\mathbf{X})}$ may be computed for every pair $(\mathbf{X},\mathbf{Y})$ from the joint distribution of the corresponding model $\scm_{do(\mathbf{X})}$.

    \item A vertical arrow $\alpha_{\mathbf{X'}}$ expresses a surjective stochastic function between the set of outcomes of low-level variable(s) $a^{-1}(\mathbf{X'})$ and the outcomes of the high-level variable(s) $\mathbf{X'}$. Again, because of the finiteness of the sets, these maps may be encoded in (column-stochastic) matrices with dimension $|\scm[a^{-1}(\mathbf{X'})]| \times |\scmi[\mathbf{X'}]|$. Intuitively, every column encodes a map from a low-level outcome of $a^{-1}(\mathbf{X'})$ to a high-level outcome of $\mathbf{X'}$. The abstraction matrix has the following key properties: (i) every column contains a single one (in order to satisfy stochasticity and functionality); (ii) every row contains at least a one (in order to satisfy surjectivity).
\end{itemize}
Thus, nodes in the diagram can be read as sets and arrows as matrices. It follows that composition of arrows amounts to matrix multiplication.

\section{Moore-Penrose Inverse} \label{app:MPInverse}

\begin{definition*}[Moore-Penrose Inverse]
    Given a matrix $\alpha \in \mathbb{C}^{N \times M}$, the Moore-Penrose inverse $\mpinv{\alpha}$ is the unique matrix in $\mathbb{C}^{M \times N}$ such that: 
    (i) $\alpha \mpinv{\alpha} \alpha = \alpha$;
    (ii) $\mpinv{\alpha} \alpha \mpinv{\alpha} = \mpinv{\alpha}$;
    (iii) $(\alpha\mpinv{\alpha})^* = \alpha\mpinv{\alpha}$;
    (iv) $(\mpinv{\alpha}\alpha)^* = \mpinv{\alpha}\alpha$.
\end{definition*}

In the definition $\alpha^*$ denotes the conjugate transpose of $\alpha$.
Among others, the Moore-Penrose inverse has the following relevant properties: 
\begin{itemize}
    \item[(M1)] Existence and uniqueness: for any abstraction matrix $\alpha_X$, we can compute $\mpinv{\alpha_X}$.
    \item[(M2)] Right inverse: given that the abstraction matrix $\alpha_X$ has linearly independent rows, $\alpha_X \mpinv{\alpha_X} = I$.
    \item[(M3)] Uniformity: given that the abstraction matrix $\alpha_X$ encodes surjective maps, $\mpinv{\alpha_X}$ implies a uniform pullback of $\alpha_X$.
\end{itemize}

Properties (M1)-(M2) are standard properties of the Moore-Penrose inverse \cite{penrose1955generalized}. We prove property (M3) in the following proposition.

\begin{proposition*}
    Given an abstraction matrix $\alpha_X$ encoding a surjective function, $\mpinv{\alpha_X}$ implies a uniform pullback of $\alpha_X$.
\end{proposition*}
\emph{Proof.} Let $\alpha_X \in \mathbb{R}^{N \times M}$ be an abstraction matrix. Recall that the abstraction matrix can be read as encoding a surjective mapping from a domain with cardinality $M$ to a codomain with cardinality $N$: a row $j$ encodes which elements of the domain are mapped onto element $j$ of the codomain.

We restrict the domain from the complex number $\mathbb{C}$ to the real number $\mathbb{R}$ because, by definition, the elements of the abstraction matrix only assume binary values. 
Moreover, by definition, we know that $N \leq M$ because of the surjectivity requirement, and that, by construction, rows are necessarily linearly independent.

Because of the linearly independent rows and the real values, the Moore-Penrose inverse assumes the following form:
\begin{equation*}
   \mpinv{\alpha_X} = \alpha_X^T (\alpha_X \alpha_X^T)^{-1},
\end{equation*}
where $\alpha_X^T$ is the transpose of $\alpha_X$. 

The matrix $\alpha_X^T$ is a binary matrix with linearly independent columns of shape $\mathbb{R}^{M \times N}$. A column $j$ encodes which elements of the domain are mapped onto element $j$ of the codomain.

Let us then consider the product $\alpha_X \alpha_X^T$. Because of the linear independence, it is immediate that, $\sum_{i=1}^M {\alpha_X}_{ji} {\alpha_X^T}_{ik} \neq 0$ iff $j=k$.
Moreover, because of the binary form, $\sum_{i=1}^M {\alpha_X}_{ji} {\alpha_X^T}_{ij} = \sum_{i=1}^M {\alpha_X}_{ji} = \sum_{i=1}^M {\alpha_X^T}_{ij}$; that is, the number of ones on the $j^{th}$ row of $\alpha_X$ or, equivalently, the number of ones on the $j^{th}$ column of $\alpha_X^T$. Thus, $\alpha_X \alpha_X^T$ returns a diagonal matrix of shape $\mathbb{R}^{N \times N}$, where the $j^{th}$ element of the diagonal gives the number of ones in the $j^{th}$ row of $\alpha_X$. Intuitively, element $(\alpha_X \alpha_X^T)_{jj}$ is the number of elements from the domain that are mapped onto the same element $j$ in the codomain.

The inverse $(\alpha_X \alpha_X^T)^{-1}$ of the diagonal matrix $\alpha_X \alpha_X^T$  is again a diagonal matrix of shape $\mathbb{R}^{N \times N}$, with inverted elements. Intuitively, element $(\alpha_X \alpha_X^T)^{-1}_{jj} = \frac{1}{(\alpha_X \alpha_X^T)_{jj}}$ simply assign a weight to each element $j$ in the codomain proportional to the number of elements in the domain that are mapped onto $j$.

Finally, the product $\alpha_X^T (\alpha_X \alpha_X^T)^{-1}$ scales the matrix $\alpha_X^T$ by the diagonal matrix of weights $(\alpha_X \alpha_X^T)^{-1}$. Because of the diagonal form of $(\alpha_X \alpha_X^T)^{-1}$, it follows that $\sum_{i=1}^N {\alpha_X^T}_{ji} {(\alpha_X \alpha_X^T)^{-1}}_{ik}$ iff $i=k$. Hence, the result is a non-binary matrix with linearly independent columns of shape $\mathbb{R}^{M \times N}$, where a column $j$ encodes which elements of the domain are mapped onto element $j$ of the codomain and sums up to $1$. This, then, encodes a uniform pullback from the codomain to the domain. $\blacksquare$

\section{Independence-based definition of a causal assessment set} \label{app:independencebaseddef}

We present here an alternative definition of a causal assessment set based on the notion of independence instead of ordering.

\begin{definition}[Independent assessment set]
Given an abstraction $\absi$ from $\scm$ to $\scmi$, let $\mathcal{J}_i$ be the set of all non-empty disjoint pairs $(\mathbf{X'},\mathbf{Y'}) \in \mathscr{P}(\mathcal{X'})\times\mathscr{P}(\mathcal{X'})$, where $\nexists Y' \subseteq \mathbf{Y'}$ such that $Y' \perp \mathbf{X'}$ in $\scm_{do(\mathbf{X'})}$. 
\end{definition}

This definition is equivalent to Definition \ref{def:causalAssessmentSet}, as it forbids independence instead of path d-separation. Further, notice that the condition in the above definition may be restated as the requirement that 
$\nexists Y'\subseteq \mathbf{Y'}$ such that $P(\mathbf{Y'} \vert do(\mathbf{X'})) = P(\mathbf{Y'}\setminus Y' \vert do(\mathbf{X'})) P(Y')$. This formulation highlights that, if the condition is not met, than a measure of error would be affected by the observational quantity $P(Y')$.

\section{Properties of Error} \label{app:errorproperties}

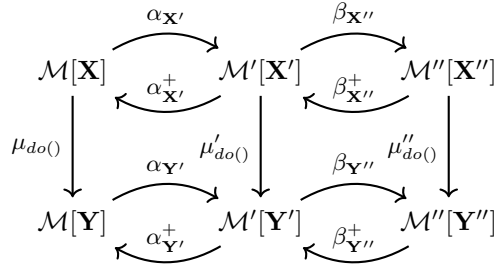
\begin{figure}
    \centering
    \begin{tikzpicture}[shorten >=1pt, auto, node distance=1cm, thick, scale=1.0, every node/.style={scale=1.0}]

\node[] (M0_0) at (0,0) {$\scm[\mathbf{X}]$};
\node[] (M0_1) at (2.5,0) {$\scmi[\mathbf{X'}]$};
\node[] (M0_2) at (5,0) {$\scmii[\mathbf{X''}]$};
\node[] (M1_0) at (0,-2) {$\scm[\mathbf{Y}]$};
\node[] (M1_1) at (2.5,-2) {$\scmi[\mathbf{Y'}]$};
\node[] (M1_2) at (5,-2) {$\scmii[\mathbf{Y''}]$};

\draw[->,bend left]  (M0_0) to node[above,font=\small]{$\alpha_\mathbf{X'}$} (M0_1);
\draw[->,bend left]  (M0_1) to node[above,font=\small]{$\beta_\mathbf{X''}$} (M0_2);

\draw[->,bend left]  (M0_1) to node[above,font=\small]{$\mpinv{\alpha_\mathbf{X'}}$} (M0_0);
\draw[->,bend left]  (M0_2) to node[above,font=\small]{$\mpinv{\beta_\mathbf{X''}}$} (M0_1);

\draw[->,bend left]  (M1_0) to node[above,font=\small]{$\alpha_\mathbf{Y'}$} (M1_1);
\draw[->,bend left]  (M1_1) to node[above,font=\small]{$\beta_\mathbf{Y''}$} (M1_2);
\draw[->,bend left]  (M1_1) to node[above,font=\small]{$\mpinv{\alpha_\mathbf{Y'}}$} (M1_0);
\draw[->,bend left]  (M1_2) to node[above,font=\small]{$\mpinv{\beta_\mathbf{Y''}}$} (M1_1);

\draw[->]  (M0_0) to node[left,font=\small]{$\mu_{do()}$} (M1_0);
\draw[->]  (M0_1) to node[left,font=\small]{$\mu'_{do()}$} (M1_1);
\draw[->]  (M0_2) to node[left,font=\small]{$\mu''_{do()}$} (M1_2);

\end{tikzpicture}
    \caption{Error diagram for abstractions $\absi$ and $\absii$. For readability, we omit $a^{-1}()$ and the $do()$ argument.}
    \label{fig:consitency_triangle}
\end{figure}

We here prove properties about our measures of error, with referenc to the diagram in Fig. \ref{fig:consitency_triangle}.

\subsection{Property (E1): Triangle Inequality} \label{app:C1}
\begin{proposition*}
Given two abstractions $\absi$ from $\scm$ to $\scmi$, and $\absii$ from $\scmi$ to $\scmii$, $\EI{\absii\absi,\mathbf{X''},\mathbf{Y''}} \leq \EI{\absi,\mathbf{X'},\mathbf{Y'}} + \EI{\absii,\mathbf{X''},\mathbf{Y''}}$.
\end{proposition*}

\emph{Proof.} We prove this property for the four interventional measures IC, IIL, ISIL and ISC. The proofs are similar, but not identical due to the different orientation of edges in the diagram of interest.

(a) Let us start with IC, defined in Diagram \ref{fig:app_IC}.
We have the following individual errors:
\begin{align*}
\EIC{\absi,\mathbf{X'},\mathbf{Y'}} & = D_{JSD}(\alpha_\mathbf{Y'} \mu, \mu' \alpha_\mathbf{X'})\\
\EIC{\absii,\mathbf{X''},\mathbf{Y''}} & = D_{JSD}(\beta_\mathbf{Y''} \mu', \mu'' \beta_\mathbf{X''}).
\end{align*}

By (J5), we have:
\begin{align*}
D_{JSD}(\alpha_\mathbf{Y'} \mu, \mu' \alpha_\mathbf{X'}) & \geq D_{JSD}(\beta_\mathbf{Y''} \alpha_\mathbf{Y'} \mu, \beta_\mathbf{Y''} \mu' \alpha_\mathbf{X'})\\
D_{JSD}(\beta_\mathbf{Y''} \mu', \mu'' \beta_\mathbf{X''}) & \geq D_{JSD}(\beta_\mathbf{Y''} \mu' \alpha_\mathbf{X'}, \mu'' \beta_\mathbf{X''} \alpha_\mathbf{X'}).
\end{align*}

By (J4), we have:
\begin{align*}
& D_{JSD}(\beta_\mathbf{Y''} \alpha_\mathbf{Y'} \mu, \beta_\mathbf{Y''} \mu' \alpha_\mathbf{X'}) \\
+ & D_{JSD}(\beta_\mathbf{Y''} \mu' \alpha_\mathbf{X'}, \mu'' \beta_\mathbf{X''} \alpha_\mathbf{X'}) \\ \geq &
D_{JSD}(\beta_\mathbf{Y''} \alpha_\mathbf{Y'} \mu,\mu'' \beta_\mathbf{X''} \alpha_\mathbf{X'}),
\end{align*}
where, by definition, $D_{JSD}(\beta_\mathbf{Y''} \alpha_\mathbf{Y'} \mu,\mu'' \beta_\mathbf{X''} \alpha_\mathbf{X'}) = \EIC{\absii\absi,\mathbf{X''},\mathbf{Y''}}$.
Hence:
$\EIC{\absii\absi,\mathbf{X''},\mathbf{Y''}} \leq \EIC{\absi,\mathbf{X'},\mathbf{Y'}} + \EIC{\absii,\mathbf{X''},\mathbf{Y''}}$.

\begin{figure}
    \centering
    \begin{tikzpicture}[shorten >=1pt, auto, node distance=1cm, thick, scale=1.0, every node/.style={scale=1.0}]

\node[] (M0_0) at (0,0) {$\scm[\mathbf{X}]$};
\node[] (M0_1) at (2.5,0) {$\scmi[\mathbf{X'}]$};
\node[] (M0_2) at (5,0) {$\scmii[\mathbf{X''}]$};
\node[] (M1_0) at (0,-2) {$\scm[\mathbf{Y}]$};
\node[] (M1_1) at (2.5,-2) {$\scmi[\mathbf{Y'}]$};
\node[] (M1_2) at (5,-2) {$\scmii[\mathbf{Y''}]$};

\draw[->]  (M0_0) to node[above,font=\small]{$\alpha_\mathbf{X'}$} (M0_1);
\draw[->]  (M0_1) to node[above,font=\small]{$\beta_\mathbf{X''}$} (M0_2);


\draw[->]  (M1_0) to node[above,font=\small]{$\alpha_\mathbf{Y'}$} (M1_1);
\draw[->]  (M1_1) to node[above,font=\small]{$\beta_\mathbf{Y''}$} (M1_2);

\draw[->]  (M0_0) to node[left,font=\small]{$\mu$} (M1_0);
\draw[->]  (M0_1) to node[left,font=\small]{$\mu'$} (M1_1);
\draw[->]  (M0_2) to node[left,font=\small]{$\mu''$} (M1_2);

\end{tikzpicture}
    \caption{Error diagram for IC}
    \label{fig:app_IC}
\end{figure}

(b) Let us consider IIL, defined in Diagram \ref{fig:app_IIL}.
We have the following individual errors:
\begin{align*}
\EIIL{\absi,\mathbf{X'},\mathbf{Y'}} & = D_{JSD}(\mu, \mpinv{\alpha_\mathbf{Y'}} \mu' \alpha_\mathbf{X'})\\
\EIIL{\absii,\mathbf{X''},\mathbf{Y''}} & = D_{JSD}(\mu', \mpinv{\beta_\mathbf{Y''}} \mu'' \beta_\mathbf{X''}).
\end{align*}

By (J5), we have:
\begin{align*}
D_{JSD}(\mu', \mpinv{\beta_\mathbf{Y''}} \mu'' \beta_\mathbf{X''}) & \geq D_{JSD}(\mpinv{\alpha_\mathbf{Y'}} \mu' \alpha_\mathbf{X'}, \mpinv{\alpha_\mathbf{Y'}} \mpinv{\beta_\mathbf{Y''}} \mu'' \beta_\mathbf{X''} \alpha_\mathbf{X'}).
\end{align*}

By (J4), we have:
\begin{align*}
& D_{JSD}(\mu, \mpinv{\alpha_\mathbf{Y'}} \mu' \alpha_\mathbf{X'}) \\
+ & D_{JSD}(\mpinv{\alpha_\mathbf{Y'}} \mu' \alpha_\mathbf{X'}, \mpinv{\alpha_\mathbf{Y'}} \mpinv{\beta_\mathbf{Y''}} \mu'' \beta_\mathbf{X''} \alpha_\mathbf{X'}) \\ \geq &
D_{JSD}(\mu,\mpinv{\alpha_\mathbf{Y'}} \mpinv{\beta_\mathbf{Y''}} \mu'' \beta_\mathbf{X''} \alpha_\mathbf{X'}),
\end{align*}
where, by definition, $D_{JSD}(\mu,\mpinv{\alpha_\mathbf{Y'}} \mpinv{\beta_\mathbf{Y''}} \mu'' \beta_\mathbf{X''} \alpha_\mathbf{X'}) = \EIIL{\absii\absi,\mathbf{X''},\mathbf{Y''}}$.
Hence:
$\EIIL{\absii\absi,\mathbf{X''},\mathbf{Y''}} \leq \EIIL{\absi,\mathbf{X'},\mathbf{Y'}} + \EIIL{\absii,\mathbf{X''},\mathbf{Y''}}$.

\begin{figure}
    \centering
    \begin{tikzpicture}[shorten >=1pt, auto, node distance=1cm, thick, scale=1.0, every node/.style={scale=1.0}]

\node[] (M0_0) at (0,0) {$\scm[\mathbf{X}]$};
\node[] (M0_1) at (2.5,0) {$\scmi[\mathbf{X'}]$};
\node[] (M0_2) at (5,0) {$\scmii[\mathbf{X''}]$};
\node[] (M1_0) at (0,-2) {$\scm[\mathbf{Y}]$};
\node[] (M1_1) at (2.5,-2) {$\scmi[\mathbf{Y'}]$};
\node[] (M1_2) at (5,-2) {$\scmii[\mathbf{Y''}]$};

\draw[->]  (M0_0) to node[above,font=\small]{$\alpha_\mathbf{X'}$} (M0_1);
\draw[->]  (M0_1) to node[above,font=\small]{$\beta_\mathbf{X''}$} (M0_2);


\draw[->]  (M1_1) to node[above,font=\small]{$\mpinv{\alpha_\mathbf{Y'}}$} (M1_0);
\draw[->]  (M1_2) to node[above,font=\small]{$\mpinv{\beta_\mathbf{Y''}}$} (M1_1);

\draw[->]  (M0_0) to node[left,font=\small]{$\mu$} (M1_0);
\draw[->]  (M0_1) to node[left,font=\small]{$\mu'$} (M1_1);
\draw[->]  (M0_2) to node[left,font=\small]{$\mu''$} (M1_2);

\end{tikzpicture}
    \caption{Error diagram for IIL}
    \label{fig:app_IIL}
\end{figure}

(c) Let us consider ISIL, defined in Diagram \ref{fig:app_ISIL}.
We have the following individual errors:
\begin{align*}
\EISIL{\absi,\mathbf{X'},\mathbf{Y'}} & = D_{JSD}(\mu', \alpha_\mathbf{Y'} \mu \mpinv{\alpha_\mathbf{X'}})\\
\EISIL{\absii,\mathbf{X''},\mathbf{Y''}} & = D_{JSD}(\mu'', \beta_\mathbf{Y''} \mu' \mpinv{\beta_\mathbf{X''}}).
\end{align*}

By (J5), we have:
\begin{align*}
D_{JSD}(\mu', \alpha_\mathbf{Y'} \mu \mpinv{\alpha_\mathbf{X'}}) & \geq D_{JSD}(\beta_\mathbf{Y''} \mu' \mpinv{\beta_\mathbf{X''}}, \beta_\mathbf{Y''} \alpha_\mathbf{Y'} \mu \mpinv{\alpha_\mathbf{X'}} \mpinv{\beta_\mathbf{X''}}).
\end{align*}

By (J4), we have:
\begin{align*}
& D_{JSD}(\mu'', \beta_\mathbf{Y''} \mu' \mpinv{\beta_\mathbf{X''}}) \\
+ & D_{JSD}(\beta_\mathbf{Y''} \mu' \mpinv{\beta_\mathbf{X''}}, \beta_\mathbf{Y''} \alpha_\mathbf{Y'} \mu \mpinv{\alpha_\mathbf{X'}} \mpinv{\beta_\mathbf{X''}}) \\ \geq &
D_{JSD}(\mu'',\beta_\mathbf{Y''} \alpha_\mathbf{Y'} \mu \mpinv{\alpha_\mathbf{X'}} \mpinv{\beta_\mathbf{X''}}),
\end{align*}
where, by definition, $D_{JSD}(\mu'',\beta_\mathbf{Y''} \alpha_\mathbf{Y'} \mu \mpinv{\alpha_\mathbf{X'}} \mpinv{\beta_\mathbf{X''}}) = \EISIL{\absii\absi,\mathbf{X''},\mathbf{Y''}}$.
Hence:
$\EISIL{\absii\absi,\mathbf{X''},\mathbf{Y''}} \leq \EISIL{\absi,\mathbf{X'},\mathbf{Y'}} + \EISIL{\absii,\mathbf{X''},\mathbf{Y''}}$.

\begin{figure}
    \centering
    \begin{tikzpicture}[shorten >=1pt, auto, node distance=1cm, thick, scale=1.0, every node/.style={scale=1.0}]

\node[] (M0_0) at (0,0) {$\scm[\mathbf{X}]$};
\node[] (M0_1) at (2.5,0) {$\scmi[\mathbf{X'}]$};
\node[] (M0_2) at (5,0) {$\scmii[\mathbf{X''}]$};
\node[] (M1_0) at (0,-2) {$\scm[\mathbf{Y}]$};
\node[] (M1_1) at (2.5,-2) {$\scmi[\mathbf{Y'}]$};
\node[] (M1_2) at (5,-2) {$\scmii[\mathbf{Y''}]$};

\draw[->]  (M0_1) to node[above,font=\small]{$\mpinv{\alpha_\mathbf{X'}}$} (M0_0);
\draw[->]  (M0_2) to node[above,font=\small]{$\mpinv{\beta_\mathbf{X''}}$} (M0_1);

\draw[->]  (M1_0) to node[above,font=\small]{$\alpha_\mathbf{Y'}$} (M1_1);
\draw[->]  (M1_1) to node[above,font=\small]{$\beta_\mathbf{Y''}$} (M1_2);

\draw[->]  (M0_0) to node[left,font=\small]{$\mu$} (M1_0);
\draw[->]  (M0_1) to node[left,font=\small]{$\mu'$} (M1_1);
\draw[->]  (M0_2) to node[left,font=\small]{$\mu''$} (M1_2);

\end{tikzpicture}
    \caption{Error diagram for ISIL}
    \label{fig:app_ISIL}
\end{figure}

(d) Let us consider ISC, defined in Diagram \ref{fig:app_ISC}.
We have the following individual errors:
\begin{align*}
\EISC{\absi,\mathbf{X'},\mathbf{Y'}} & = D_{JSD}(\mu \mpinv{\alpha_\mathbf{X'}}, \mpinv{\alpha_\mathbf{Y'}} \mu')\\
\EISC{\absii,\mathbf{X''},\mathbf{Y''}} & = D_{JSD}(\mu' \mpinv{\beta_\mathbf{X''}}, \mpinv{\beta_\mathbf{Y''}} \mu'').
\end{align*}

By (J5), we have:
\begin{align*}
D_{JSD}(\mu \mpinv{\alpha_\mathbf{X'}}, \mpinv{\alpha_\mathbf{Y'}} \mu') & \geq D_{JSD}(\mu \mpinv{\alpha_\mathbf{X'}} \mpinv{\beta_\mathbf{X''}}, \mpinv{\alpha_\mathbf{Y'}} \mu' \mpinv{\beta_\mathbf{X''}}) \\
D_{JSD}(\mu' \mpinv{\beta_\mathbf{X''}}, \mpinv{\beta_\mathbf{Y''}} \mu'') & \geq
D_{JSD}(\mpinv{\alpha_\mathbf{Y'}} \mu' \mpinv{\beta_\mathbf{X''}}, \mpinv{\alpha_\mathbf{Y'}} \mpinv{\beta_\mathbf{Y''}} \mu'').
\end{align*}

By (J4), we have:
\begin{align*}
& D_{JSD}(\mu \mpinv{\alpha_\mathbf{X'}} \mpinv{\beta_\mathbf{X''}}, \mpinv{\alpha_\mathbf{Y'}} \mu' \mpinv{\beta_\mathbf{X''}}) \\
+ & D_{JSD}(\mpinv{\alpha_\mathbf{Y'}} \mu' \mpinv{\beta_\mathbf{X''}}, \mpinv{\alpha_\mathbf{Y'}} \mpinv{\beta_\mathbf{Y''}} \mu'') \\ \geq &
D_{JSD}(\mu \mpinv{\alpha_\mathbf{X'}} \mpinv{\beta_\mathbf{X''}},\mpinv{\alpha_\mathbf{Y'}} \mpinv{\beta_\mathbf{Y''}} \mu''),
\end{align*}
where, by definition, $D_{JSD}(\mu \mpinv{\alpha_\mathbf{X'}} \mpinv{\beta_\mathbf{X''}},\mpinv{\alpha_\mathbf{Y'}} \mpinv{\beta_\mathbf{Y''}} \mu'') = \EISC{\absii\absi,\mathbf{X''},\mathbf{Y''}}$.
Hence:
$\EISC{\absii\absi,\mathbf{X''},\mathbf{Y''}} \leq \EISC{\absi,\mathbf{X'},\mathbf{Y'}} + \EISC{\absii,\mathbf{X''},\mathbf{Y''}}$. $\blacksquare$

\begin{figure}
    \centering
    \begin{tikzpicture}[shorten >=1pt, auto, node distance=1cm, thick, scale=1.0, every node/.style={scale=1.0}]

\node[] (M0_0) at (0,0) {$\scm[\mathbf{X}]$};
\node[] (M0_1) at (2.5,0) {$\scmi[\mathbf{X'}]$};
\node[] (M0_2) at (5,0) {$\scmii[\mathbf{X''}]$};
\node[] (M1_0) at (0,-2) {$\scm[\mathbf{Y}]$};
\node[] (M1_1) at (2.5,-2) {$\scmi[\mathbf{Y'}]$};
\node[] (M1_2) at (5,-2) {$\scmii[\mathbf{Y''}]$};

\draw[->]  (M0_1) to node[above,font=\small]{$\mpinv{\alpha_\mathbf{X'}}$} (M0_0);
\draw[->]  (M0_2) to node[above,font=\small]{$\mpinv{\beta_\mathbf{X''}}$} (M0_1);

\draw[->]  (M1_1) to node[above,font=\small]{$\mpinv{\alpha_\mathbf{Y'}}$} (M1_0);
\draw[->]  (M1_2) to node[above,font=\small]{$\mpinv{\beta_\mathbf{Y''}}$} (M1_1);

\draw[->]  (M0_0) to node[left,font=\small]{$\mu$} (M1_0);
\draw[->]  (M0_1) to node[left,font=\small]{$\mu'$} (M1_1);
\draw[->]  (M0_2) to node[left,font=\small]{$\mu''$} (M1_2);

\end{tikzpicture}
    \caption{Error diagram for ISC}
    \label{fig:app_ISC}
\end{figure}

\subsection{Property (E2): Non-monotonicity}
\begin{proposition*}
Given two abstractions $\absi$ from $\scm$ to $\scmi$, and $\absii$ from $\scmi$ to $\scmii$, it is not guaranteed that $\EI{\absii\absi,\mathbf{X''},\mathbf{Y''}} \geq \EI{\absi,\mathbf{X'},\mathbf{Y'}}$.
\end{proposition*}

\emph{Proof.} We will prove this statement by counterexample, showing at once that monotonicity of IC, IIL, ISIL and ISC is not guaranteed.

Consider any one of the error diagrams in Fig. \ref{fig:app_IC}-\ref{fig:app_ISC}. Assume all the domains ($\scm[\mathbf{X}], \scmi[\mathbf{X'}], \scmii[\mathbf{X''}], \scm[\mathbf{Y}], \scmi[\mathbf{Y'}], \scmii[\mathbf{Y''}]$) to be binary and all abstraction matrices $\alpha_{\mathbf{X'}}, \alpha_{\mathbf{Y'}}, \beta_{\mathbf{X'}}, \beta_\mathbf{Y'}$ to be identity matrices $I$. Further, assume $\mu = \mu''$, but $\mu \neq \mu'$.
Because of the difference in mechanism, it is immediate to show that, by (J1) and (J2), it holds that:
$$
\EI{\absi,\mathbf{X'},\mathbf{Y'}} > 0.
$$
Similarly, by the identity of mechanism, and by (J2), it is immediate to show that:
$$
\EI{\absii\absi,\mathbf{X''},\mathbf{Y''}} = 0.
$$
Hence, $\EI{\absii\absi,\mathbf{X''},\mathbf{Y''}} < \EI{\absi,\mathbf{X'},\mathbf{Y'}}$. $\blacksquare$

\subsection{Property (NE1): Vertical non-compositionality}
\begin{proposition*}
Given an abstractions $\absi$ from $\scm$ to $\scmi$, if $\EI{\absi,\mathbf{X'},\mathbf{Y'}}=0$ and $\EI{\absi,\mathbf{Y'},\mathbf{Z'}}=0$, it does not follow $\EI{\absi,\mathbf{X'},\mathbf{Z'}}=0$.
\end{proposition*}
\emph{Proof.} We will prove this statement by counterexample, showing at once that vertical composition for IC, IIL, ISIL and ISC is not guaranteed.

Consider the following base model $\scm$:
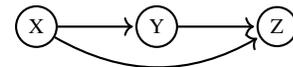
\begin{center}
\begin{tikzpicture}[shorten >=1pt, auto, node distance=1cm, thick, scale=0.8, every node/.style={scale=0.8}]
	\tikzstyle{node_style} = [circle,draw=black]
	\node[node_style] (X) at (0,0) {X};
	\node[node_style] (W) at (2,0) {Y};
	\node[node_style] (Y) at (4,0) {Z};
	
	\draw[->]  (X) to (W);
	\draw[->]  (W) to (Y);
	\draw[->, bend right]  (X) to (Y);
\end{tikzpicture}
\end{center}
and the abstracted model $\scmi$:
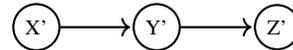
\begin{center}
\begin{tikzpicture}[shorten >=1pt, auto, node distance=1cm, thick, scale=0.8, every node/.style={scale=0.8}]
	\tikzstyle{node_style} = [circle,draw=black]
	\node[node_style] (X) at (0,0) {X'};
	\node[node_style] (W) at (2,0) {Y'};
	\node[node_style] (Y) at (4,0) {Z'};
	
	\draw[->]  (X) to (W);
	\draw[->]  (W) to (Y);
\end{tikzpicture}
\end{center}
Let us define the abstraction $\absi$ as $R=\{X,Y,Z\}$, $a=\{X\mapsto X',Y\mapsto Y', Z\mapsto Z'\}$, and simple identity maps for $\alpha_{X'},\alpha_{Y'},\alpha_{Z'}$. 

In order to compute the interventional error, we setup the following error diagram: 

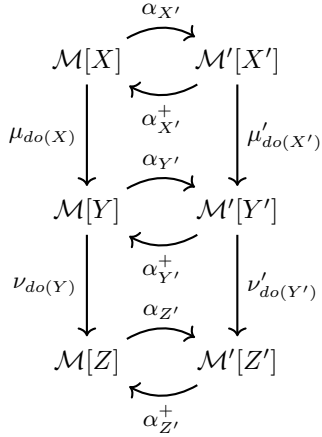
\begin{center}
\begin{tikzpicture}[shorten >=1pt, auto, node distance=1cm, thick, scale=1.0, every node/.style={scale=1.0}]
\tikzstyle{node_style} = []

\node[node_style] (M0_0) at (0,0) {$\scm[X]$};
\node[node_style] (M0_1) at (2,0) {$\scmi[X']$};
\node[node_style] (M1_0) at (0,-2) {$\scm[Y]$};
\node[node_style] (M1_1) at (2,-2) {$\scmi[Y']$};
\node[node_style] (M2_0) at (0,-4) {$\scm[Z]$};
\node[node_style] (M2_1) at (2,-4) {$\scmi[Z']$};

\draw[->,bend left]  (M0_0) to node[above,font=\small]{$\alpha_{X'}$} (M0_1);
\draw[->,bend left]  (M1_0) to node[above,font=\small]{$\alpha_{Y'}$} (M1_1);
\draw[->,bend left]  (M2_0) to node[above,font=\small]{$\alpha_{Z'}$} (M2_1);
\draw[->,bend left]  (M0_1) to node[below,font=\small]{$\mpinv{\alpha_{X'}}$} (M0_0);
\draw[->,bend left]  (M1_1) to node[below,font=\small]{$\mpinv{\alpha_{Y'}}$} (M1_0);
\draw[->,bend left]  (M2_1) to node[below,font=\small]{$\mpinv{\alpha_{Z'}}$} (M2_0);

\draw[->]  (M0_0) to node[left,font=\small]{$\mu_{do(X)}$} (M1_0);
\draw[->]  (M0_1) to node[right,font=\small]{$\mu'_{do(X')}$} (M1_1);

\draw[->]  (M1_0) to node[left,font=\small]{$\nu_{do(Y)}$} (M2_0);
\draw[->]  (M1_1) to node[right,font=\small]{$\nu'_{do(Y')}$} (M2_1);
\end{tikzpicture}
\end{center}

Notice that this diagram can be seen as a vertical extension of the diagram in Fig. \ref{fig:consitency_triangle}.

Assume $\EI{\absi,X',Y'}=0$ and $\EI{\absi,Y',Z'}=0$. Because the $\alpha$ maps are identities, this implies $\mu_{do(X)}= \mu'_{do(X')}$ and $\nu_{do(Y)}= \nu'_{do(Y')}$. Notice, in particular, how the abstracted stochastic matrix $\nu'_{do(Y')}$ would have to account for the noise introduce by the edge $X \rightarrow Z$ in the base model.

Now, if we consider the composed error $\EI{\absi,X'',Z''}$, we would obtain two different stochastic functions. In the simple abstracted model we would obtain that $\nu' \circ \mu'$ is equal to the matrix product $\nu'_{do(Y')} \mu'_{do(X')}$, yielding the composition of two edges. In the base model, instead, because of the contribution of the direct edge $X \rightarrow Z$, the stochastic function accounting for the interventional distribution on $Z$ would not necessarily be the same as the matrix product $\nu_{do(Y)} \mu_{do(X)}$. Hence: $\EI{\absi,\mathbf{X''},\mathbf{Z''}} \neq 0$.  $\blacksquare$

As a remark, it may be important to underline that stating that vertical compositionality does not hold, it does not mean that compositionality in the underlying category does not hold. Compositionality in $\mathtt{FinStoch}$ holds by definition, as it clearly exists a morphism given by the matrix product $\nu_{do(Y)} \mu_{do(X)}$. What does not hold is a naive compositionality derived from the diagram above; such compositionality does not hold because stochastic matrices are actually computed from different post-interventional models and information on how these stochastic functions are related is lost on the diagram.

\subsection{Property (NE2): Product non-compositionality}

\begin{proposition*}
Given an abstractions $\absi$ from $\scm$ to $\scmi$, if $\EI{\absi,\mathbf{X'},\mathbf{Z'}}=0$ and $\EI{\absi,\mathbf{Y'},\mathbf{Z'}}=0$, it does not follow $\EI{\absi,\mathbf{X'} \times \mathbf{Y'},\mathbf{Z'}}=0$.
\end{proposition*}
\emph{Proof.} We will prove this statement by counterexample, showing at once that monoidal composition for IC, IIL, ISIL and ISC is not guaranteed.

Consider the following base model $\scm$:
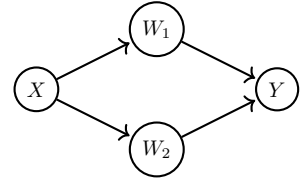
\begin{center}
\begin{tikzpicture}[shorten >=1pt, auto, node distance=1cm, thick, scale=0.8, every node/.style={scale=0.8}]
	\tikzstyle{node_style} = [circle,draw=black]
	\node[node_style] (X) at (-2,1) {$X$};
	\node[node_style] (W1) at (0,2) {$W_1$};
        \node[node_style] (W2) at (0,0) {$W_2$};
	\node[node_style] (Y) at (2,1) {$Y$};
	
	\draw[->]  (X) to (W1);
	\draw[->]  (X) to (W2);
	\draw[->]  (W1) to (Y);
        \draw[->]  (W2) to (Y);
\end{tikzpicture}
\end{center}
and the abstracted model $\scmi$:
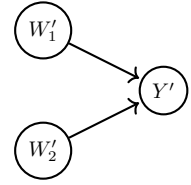
\begin{center}
\begin{tikzpicture}[shorten >=1pt, auto, node distance=1cm, thick, scale=0.8, every node/.style={scale=0.8}]
	\tikzstyle{node_style} = [circle,draw=black]
	\node[node_style] (W2) at (0,0) {$W'_2$};
	\node[node_style] (W1) at (0,2) {$W'_1$};
	\node[node_style] (Y) at (2,1) {$Y'$};
	
	\draw[->]  (W1) to (Y);
	\draw[->]  (W2) to (Y);
\end{tikzpicture}
\end{center}

Let us define the abstraction $\absi$ as $R=\{W_1,W_2,Y\}$, $a=\{W_1\mapsto W'_1,W_2\mapsto W'_2, Y\mapsto Y'\}$, and simple identity maps for $\alpha_{W'_1},\alpha_{W'_2},\alpha_{Y'}$. 

In order to compute the interventional error, we setup the following error diagrams for $\EI{\absi,W'_1,Y'}$ and $\EI{\absi,W'_2,Y'}$
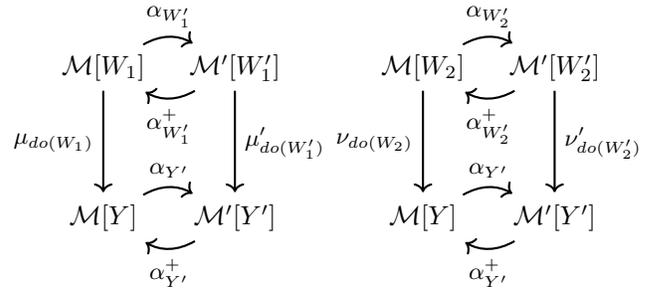
\begin{center}
\begin{tikzpicture}[shorten >=1pt, auto, node distance=1cm, thick, scale=1.0, every node/.style={scale=1.0}]
\tikzstyle{node_style} = []

\node[node_style] (M0_0) at (0,0) {$\scm[W_1]$};
\node[node_style] (M0_1) at (1.75,0) {$\scmi[W'_1]$};
\node[node_style] (M1_0) at (0,-2) {$\scm[Y]$};
\node[node_style] (M1_1) at (1.75,-2) {$\scmi[Y']$};

\node[node_style] (M2_0) at (4.25,0) {$\scm[W_2]$};
\node[node_style] (M2_1) at (6,0) {$\scmi[W'_2]$};
\node[node_style] (M3_0) at (4.25,-2) {$\scm[Y]$};
\node[node_style] (M3_1) at (6,-2) {$\scmi[Y']$};

\draw[->,bend left]  (M0_0) to node[above,font=\small]{$\alpha_{W'_1}$} (M0_1);
\draw[->,bend left]  (M1_0) to node[above,font=\small]{$\alpha_{Y'}$} (M1_1);
\draw[->,bend left]  (M0_1) to node[below,font=\small]{$\mpinv{\alpha_{W'_1}}$} (M0_0);
\draw[->,bend left]  (M1_1) to node[below,font=\small]{$\mpinv{\alpha_{Y'}}$} (M1_0);

\draw[->,bend left]  (M2_0) to node[above,font=\small]{$\alpha_{W'_2}$} (M2_1);
\draw[->,bend left]  (M3_0) to node[above,font=\small]{$\alpha_{Y'}$} (M3_1);
\draw[->,bend left]  (M2_1) to node[below,font=\small]{$\mpinv{\alpha_{W'_2}}$} (M2_0);
\draw[->,bend left]  (M3_1) to node[below,font=\small]{$\mpinv{\alpha_{Y'}}$} (M3_0);

\draw[->]  (M0_0) to node[left,font=\small]{$\mu_{do(W_1)}$} (M1_0);
\draw[->]  (M0_1) to node[right,font=\small]{$\mu'_{do(W'_1)}$} (M1_1);

\draw[->]  (M2_0) to node[left,font=\small]{$\nu_{do(W_2)}$} (M3_0);
\draw[->]  (M2_1) to node[right,font=\small]{$\nu'_{do(W'_2)}$} (M3_1);

\end{tikzpicture}
\end{center}

Assume $\EI{\absi,W'_1,Y'}=0$ and $\EI{\absi,W'_2,Y'}=0$. Because the $\alpha$ maps are identity, this implies $\mu_{do(W_1)}= \mu'_{do(W'_1)}$ and $\nu_{do(W_2)}= \nu'_{do(W'_2)}$. To guarantee the identity, the abstracted stochastic matrix $\mu'_{do(W'_1)}$ and $\nu'_{do(W'_2)}$ have to account for the noise introduce by the variable $X$ in the base model.

Let us now consider the error diagram for $\EI{\absi,W'_1\times W'_2,Y'}$:

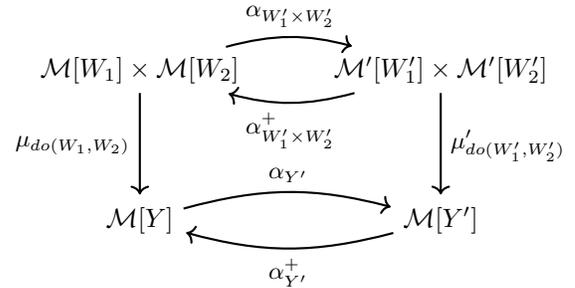
\begin{center}
\begin{tikzpicture}[shorten >=1pt, auto, node distance=1cm, thick, scale=1.0, every node/.style={scale=1.0}]
\tikzstyle{node_style} = []

\node[node_style] (M0_0) at (0,0) {$\scm[W_1] \times \scm[W_2]$};
\node[node_style] (M0_1) at (4,0) {$\scmi[W'_1] \times \scmi[W'_2]$};
\node[node_style] (M1_0) at (0,-2) {$\scm[Y]$};
\node[node_style] (M1_1) at (4,-2) {$\scm[Y']$};

\draw[->,bend left=15]  (M0_0) to node[above,font=\small]{$\alpha_{W'_1\times W'_2}$} (M0_1);
\draw[->,bend left=15]  (M1_0) to node[above,font=\small]{$\alpha_{Y'}$} (M1_1);
\draw[->,bend left=15]  (M0_1) to node[below,font=\small]{$\mpinv{\alpha_{W'_1 \times W'_2}}$} (M0_0);
\draw[->,bend left=15]  (M1_1) to node[below,font=\small]{$\mpinv{\alpha_{Y'}}$} (M1_0);

\draw[->]  (M0_0) to node[left,font=\small]{$\mu_{do(W_1,W_2)}$} (M1_0);
\draw[->]  (M0_1) to node[right,font=\small]{$\mu'_{do(W'_1,W'_2)}$} (M1_1);

\end{tikzpicture}
\end{center}

Now, the stochastic functions defined in the base and the abstracted model differs. In the base model, the stochastic function $\mu_{do(W_1,W_2)}$ is independent from the noise introduced through variable $X$, since this variable is separated in the post-interventional model $\scm_{do(W_1,W_2)}$. However, as showed above, the contribution of the noise introduced by $X$ is integrated in the stochastic functions of the abstracted model; intervening on $W'_1$ and $W'_2$ can not remove this stochastic contribution integrated in $\mu'_{do(W'_1,W'_2)}$. Hence, even if $\EI{\absi,\mathbf{X'},\mathbf{Z'}}=0$ and $\EI{\absi,\mathbf{Y'},\mathbf{Z'}}=0$, $\EI{\absi,\mathbf{X'} \times \mathbf{Y'},\mathbf{Z'}}\neq0$. $\blacksquare$

As in the case of (NE1), this property does not mean that proper monoidal compositionality in the underlying category does not hold. What does not hold is naive compositionality derived from the diagrams above; again, this compositionality does not hold because stochastic matrices are actually computed from different post-interventional models and information on how these stochastic functions are related is lost on the diagram.

\subsection{Property (E3): Zero at identity}
\begin{proposition*}
Given an abstraction $\absi$ from  $\scm$ to $\scmi$, if $\absi$ is an identity, then  $\EI{\absi,\mathbf{X'},\mathbf{Y'}}=0$.
\end{proposition*}

\emph{Proof.}
Trivially, with reference to Fig. \ref{fig:consitency_triangle}, if $\mu=\mu'$ and $\alpha_{\mathbf{X'}}$ and $\alpha_{\mathbf{Y'}}$ are identities, then any $D_{JSD}(\cdot, \cdot)$ of interest on the diagram is $0$. Hence, $\EI{\absi,\mathbf{X'},\mathbf{Y'}}=0$. $\blacksquare$

\subsection{Property (E4): Zero at singleton}
\begin{proposition*}
Given an abstraction $\absi$ from  $\scm$ to $\scmi$, if $\scmi$ is a singleton, then $\EIC{\absi,\mathbf{X'},\mathbf{Y'}}=0$.
\end{proposition*}

\emph{Proof.} If $\scmi$ is a singleton, then the error diagram reduces to:

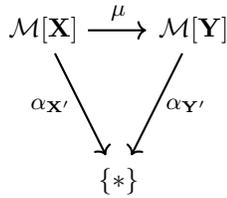
\begin{center}
\begin{tikzpicture}[shorten >=1pt, auto, node distance=1cm, thick, scale=1.0, every node/.style={scale=1.0}]
\tikzstyle{node_style} = []

\node[node_style] (M0_0) at (0,0) {$\scm[\mathbf{X}]$};
\node[node_style] (M0_1) at (2,0) {$\scm[\mathbf{Y}]$};
\node[node_style] (M1_0) at (1,-2) {$\{*\}$};

\draw[->]  (M0_0) to node[above,font=\small]{$\mu$} (M0_1);

\draw[->]  (M0_0) to node[left,font=\small]{$\alpha_{\mathbf{X}'}$} (M1_0);
\draw[->]  (M0_1) to node[right,font=\small]{$\alpha_{\mathbf{Y}'}$} (M1_0);
\end{tikzpicture}
\end{center}

If we consider the distance
$D_{JSD}(\alpha_{\mathbf{X}'}, \alpha_\mathbf{Y'} \mu)$, this necessarily translate into the distance between two distributions putting all mass probability on the only element $*$ of the singleton. Hence $D_{JSD}([1], [1])$. 
By (J2), then, $\EIC{\absi,\mathbf{X'},\mathbf{Y'}}=0$. $\blacksquare$

\setcounter{proposition}{0}
\subsection{Proposition 1}
\begin{proposition}[Relationship between measures]
We have a partial ordering among the interventional measures as:
$\EIIL{\absi,\mathbf{X'},\mathbf{Y'}} \geq \EIC{\absi,\mathbf{X'},\mathbf{Y'}}$, 
$\EIIL{\absi,\mathbf{X'},\mathbf{Y'}} \geq \EISC{\absi,\mathbf{X'},\mathbf{Y'}}$,
$\EIC{\absi,\mathbf{X'},\mathbf{Y'}} \geq \EISIL{\absi,\mathbf{X'},\mathbf{Y'}}$,
$\EISC{\absi,\mathbf{X'},\mathbf{Y'}} \geq \EISIL{\absi,\mathbf{X'},\mathbf{Y'}}$.
\end{proposition}

\emph{Proof.} All relations can be proved by pre- or post-processing according to (J5) and applying the Moore-Penrose property (M2) to the definition of each measure.

(i) Let us start with $\EIIL{\absi,\mathbf{X'},\mathbf{Y'}} \geq \EIC{\absi,\mathbf{X'},\mathbf{Y'}}$. With reference to the diagram in Fig. \ref{fig:consitency_triangle}, we have:
\begin{equation*}
    \EIIL{\absi,\mathbf{X'},\mathbf{Y'}} = D_{JSD}( \mu, \mpinv{\alpha_\mathbf{Y'}} \mu' \alpha_\mathbf{X'} ).
\end{equation*}

By pre-processing according to (J5), we have:
\begin{equation*}
    D_{JSD}( \mu, \mpinv{\alpha_\mathbf{Y'}} \mu' \alpha_\mathbf{X'} ) \geq D_{JSD}( \alpha_\mathbf{Y'} \mu, \alpha_\mathbf{Y'} \mpinv{\alpha_\mathbf{Y'}} \mu' \alpha_\mathbf{X'} ),
\end{equation*}

which, by the Moore-Penrose property (M2), simplifies to:
\begin{equation*}
    D_{JSD}( \mu, \mpinv{\alpha_\mathbf{Y'}} \mu' \alpha_\mathbf{X'} ) \geq D_{JSD}( \alpha_\mathbf{Y'} \mu, \mu' \alpha_\mathbf{X'} ),
\end{equation*}

and, finally, by symmetry (J3):
\begin{equation*}
    D_{JSD}( \mu, \mpinv{\alpha_\mathbf{Y'}} \mu' \alpha_\mathbf{X'} ) \geq D_{JSD}( \mu' \alpha_\mathbf{X'}, \alpha_\mathbf{Y'} \mu ).
\end{equation*}

where, the rhs is $\EIC{\absi,\mathbf{X'},\mathbf{Y'}}$. Hence: $\EIIL{\absi,\mathbf{X'},\mathbf{Y'}} \geq \EIC{\absi,\mathbf{X'},\mathbf{Y'}}$.

As a remark, notice that an analogous proof aimed at showing the reverse, $\EIC{\absi,\mathbf{X'},\mathbf{Y'}} \geq \EIIL{\absi,\mathbf{X'},\mathbf{Y'}}$, would not hold because property (M2) is not symmetric. The asymmetry makes intuitive sense: it is stricter to require no loss of information than to require consistency.

(ii) Let us consider $\EIIL{\absi,\mathbf{X'},\mathbf{Y'}} \geq \EISC{\absi,\mathbf{X'},\mathbf{Y'}}$. With reference to the diagram in Fig. \ref{fig:consitency_triangle}, we have:
\begin{equation*}
    \EIIL{\absi,\mathbf{X'},\mathbf{Y'}} = D_{JSD}( \mu, \mpinv{\alpha_\mathbf{Y'}} \mu' \alpha_\mathbf{X'} ).
\end{equation*}

By post-processing according to (J5), we have:
\begin{equation*}
    D_{JSD}( \mu, \mpinv{\alpha_\mathbf{Y'}} \mu' \alpha_\mathbf{X'} ) \geq D_{JSD}( \mu \mpinv{\alpha_\mathbf{X'}}, \mpinv{\alpha_\mathbf{Y'}} \mu' \alpha_\mathbf{X'} \mpinv{\alpha_\mathbf{X'}}),
\end{equation*}

which, by the Moore-Penrose property (M2), simplifies to:
\begin{equation*}
    D_{JSD}( \mu, \mpinv{\alpha_\mathbf{Y'}} \mu' \alpha_\mathbf{X'} ) \geq D_{JSD}( \mu \mpinv{\alpha_\mathbf{X'}}, \mpinv{\alpha_\mathbf{Y'}} \mu' ),
\end{equation*}

and, finally, by symmetry (J3):
\begin{equation*}
    D_{JSD}( \mu, \mpinv{\alpha_\mathbf{Y'}} \mu' \alpha_\mathbf{X'} ) \geq D_{JSD}( \mpinv{\alpha_\mathbf{Y'}} \mu', \mu \mpinv{\alpha_\mathbf{X'}} ).
\end{equation*}

where, the rhs is $\EISC{\absi,\mathbf{X'},\mathbf{Y'}}$. Hence: $\EIIL{\absi,\mathbf{X'},\mathbf{Y'}} \geq \EISC{\absi,\mathbf{X'},\mathbf{Y'}}$.

(iii) Let us consider $\EIC{\absi,\mathbf{X'},\mathbf{Y'}} \geq \EISIL{\absi,\mathbf{X'},\mathbf{Y'}}$. With reference to the diagram in Fig. \ref{fig:consitency_triangle}, we have:
\begin{equation*}
    \EIC{\absi,\mathbf{X'},\mathbf{Y'}} = D_{JSD}( \alpha_\mathbf{Y'}\mu, \mu' \alpha_\mathbf{X'} ).
\end{equation*}

By post-processing according to (J5), we have:
\begin{equation*}
    D_{JSD}( \alpha_\mathbf{Y'}\mu, \mu' \alpha_\mathbf{X'} ) \geq D_{JSD}( \alpha_\mathbf{Y'}\mu \mpinv{\alpha_\mathbf{X'}}, \mu' \alpha_\mathbf{X'} \mpinv{\alpha_\mathbf{X'}}),
\end{equation*}

which, by the Moore-Penrose property (M2), simplifies to:
\begin{equation*}
    D_{JSD}( \alpha_\mathbf{Y'}\mu, \mu' \alpha_\mathbf{X'} ) \geq D_{JSD}( \alpha_\mathbf{Y'}\mu \mpinv{\alpha_\mathbf{X'}}, \mu'),
\end{equation*}

and, finally, by symmetry (J3):
\begin{equation*}
    D_{JSD}( \alpha_\mathbf{Y'}\mu, \mu' \alpha_\mathbf{X'} ) \geq D_{JSD}( \mu', \alpha_\mathbf{Y'}\mu \mpinv{\alpha_\mathbf{X'}}).
\end{equation*}

where, the rhs is $\EISIL{\absi,\mathbf{X'},\mathbf{Y'}}$. Hence: $\EIC{\absi,\mathbf{X'},\mathbf{Y'}} \geq \EISIL{\absi,\mathbf{X'},\mathbf{Y'}}$.

(iv) Let us consider $\EISC{\absi,\mathbf{X'},\mathbf{Y'}} \geq \EISIL{\absi,\mathbf{X'},\mathbf{Y'}}$. With reference to the diagram in Fig. \ref{fig:consitency_triangle}, we have:
\begin{equation*}
    \EISC{\absi,\mathbf{X'},\mathbf{Y'}} = D_{JSD}( \mu \mpinv{\alpha_\mathbf{X'}}, \mpinv{\alpha_\mathbf{Y'}} \mu').
\end{equation*}

By pre-processing according to (J5), we have:
\begin{equation*}
    D_{JSD}( \mu \mpinv{\alpha_\mathbf{X'}}, \mpinv{\alpha_\mathbf{Y'}} \mu') \geq D_{JSD}( \alpha_\mathbf{Y'} \mu \mpinv{\alpha_\mathbf{X'}}, \alpha_\mathbf{Y'} \mpinv{\alpha_\mathbf{Y'}} \mu'),
\end{equation*}

which, by the Moore-Penrose property (M2), simplifies to:
\begin{equation*}
    D_{JSD}( \mu \mpinv{\alpha_\mathbf{X'}}, \mpinv{\alpha_\mathbf{Y'}} \mu') \geq D_{JSD}( \alpha_\mathbf{Y'} \mu \mpinv{\alpha_\mathbf{X'}}, \mu'),
\end{equation*}

and, finally, by symmetry (J3):
\begin{equation*}
    D_{JSD}( \mu \mpinv{\alpha_\mathbf{X'}}, \mpinv{\alpha_\mathbf{Y'}} \mu') \geq D_{JSD}( \mu', \alpha_\mathbf{Y'}\mu \mpinv{\alpha_\mathbf{X'}}).
\end{equation*}

where, the rhs is $\EISIL{\absi,\mathbf{X'},\mathbf{Y'}}$. Hence: $\EISC{\absi,\mathbf{X'},\mathbf{Y'}} \geq \EISIL{\absi,\mathbf{X'},\mathbf{Y'}}$.

(v) By partial ordering, we also have that $\EIIL{\absi,\mathbf{X'},\mathbf{Y'}} \geq \EISIL{\absi,\mathbf{X'},\mathbf{Y'}}$. With reference to the diagram in Fig. \ref{fig:consitency_triangle}, we have:
\begin{equation*}
    \EIIL{\absi,\mathbf{X'},\mathbf{Y'}} = D_{JSD}( \mu, \mpinv{\alpha_\mathbf{Y'}} \mu' \alpha_\mathbf{X'} ).
\end{equation*}

By pre- and post-processing according to (J5), we have:
\begin{equation*}
    D_{JSD}( \mu, \mpinv{\alpha_\mathbf{Y'}} \mu' \alpha_\mathbf{X'} ) \geq D_{JSD}( \alpha_\mathbf{Y'} \mu \mpinv{\alpha_\mathbf{X'}}, \alpha_\mathbf{Y'} \mpinv{\alpha_\mathbf{Y'}} \mu' \alpha_\mathbf{X'} \mpinv{\alpha_\mathbf{X'}} ),
\end{equation*}

which, by the Moore-Penrose property (M2), simplifies to:
\begin{equation*}
    D_{JSD}( \mu, \mpinv{\alpha_\mathbf{Y'}} \mu' \alpha_\mathbf{X'} ) \geq D_{JSD}( \alpha_\mathbf{Y'} \mu \mpinv{\alpha_\mathbf{X'}}, \mu'),
\end{equation*}

and, finally, by symmetry (J3):
\begin{equation*}
    D_{JSD}( \mu, \mpinv{\alpha_\mathbf{Y'}} \mu' \alpha_\mathbf{X'} ) \geq D_{JSD}( \mu', \alpha_\mathbf{Y'}\mu \mpinv{\alpha_\mathbf{X'}}).
\end{equation*}

where, the rhs is $\EISIL{\absi,\mathbf{X'},\mathbf{Y'}}$. Hence: $\EIIL{\absi,\mathbf{X'},\mathbf{Y'}} \geq \EISIL{\absi,\mathbf{X'},\mathbf{Y'}}$. $\blacksquare$

\subsection{Identity Proposition}
\begin{proposition*}[Identities]
Given an abstraction $\absi$ from  $\scm$ to $\scmi$, if $\alpha_{\mathbf{X}'} = I$, then $\EIC{\absi,\mathbf{X'},\mathbf{Y'}} = \EISIL{\absi,\mathbf{X'},\mathbf{Y'}}$ and $\EIIL{\absi,\mathbf{X'},\mathbf{Y'}} = \EISC{\absi,\mathbf{X'},\mathbf{Y'}}$; if $\alpha_{\mathbf{Y}'} = I$, then $\EIC{\absi,\mathbf{X'},\mathbf{Y'}} = \EIIL{\absi,\mathbf{X'},\mathbf{Y'}}$ and $\EISIL{\absi,\mathbf{X'},\mathbf{Y'}} = \EISC{\absi,\mathbf{X'},\mathbf{Y'}}$
\end{proposition*}

\emph{Proof.} Let us start considering the case in which $\alpha_{\mathbf{X}'} = I$. By definition of the Moore-Penrose inverse, it holds that $\alpha_\mathbf{X'} = \mpinv{\alpha_\mathbf{X'}}$. Therefore IC and ISIL simplifies as:
\begin{eqnarray*}
\EIC{\absi,\mathbf{X'},\mathbf{Y'}} = D_{JSD}(\alpha_\mathbf{Y'} \mu, \mu' \alpha_\mathbf{X'}) = D_{JSD}(\alpha_\mathbf{Y'} \mu, \mu'),\\
\EISIL{\absi,\mathbf{X'},\mathbf{Y'}} = D_{JSD}(\mu', \alpha_\mathbf{Y'} \mu \mpinv{\alpha_\mathbf{X'}}) = D_{JSD}(\mu', \alpha_\mathbf{Y'} \mu).
\end{eqnarray*}
By symmetry (J3), $\EIC{\absi,\mathbf{X'},\mathbf{Y'}} = \EISIL{\absi,\mathbf{X'},\mathbf{Y'}}$.

Moreover, IIL and ISC simplifies as:
\begin{eqnarray*}
\EIIL{\absi,\mathbf{X'},\mathbf{Y'}} = D_{JSD}( \mu, \mpinv{\alpha_\mathbf{Y'}} \mu' \alpha_\mathbf{X'} ) =  D_{JSD}( \mu, \mpinv{\alpha_\mathbf{Y'}} \mu' ),\\
\EISC{\absi,\mathbf{X'},\mathbf{Y'}} = D_{JSD}( \mu \mpinv{\alpha_\mathbf{X'}}, \mpinv{\alpha_\mathbf{Y'}} \mu') = D_{JSD}( \mu, \mpinv{\alpha_\mathbf{Y'}} \mu').
\end{eqnarray*}
Hence, $\EIIL{\absi,\mathbf{X'},\mathbf{Y'}} = \EISC{\absi,\mathbf{X'},\mathbf{Y'}}$.

Conversely, let us consider the case in which $\alpha_{\mathbf{Y}'} = I$. By definition of the Moore-Penrose inverse, it holds that $\alpha_\mathbf{Y'} = \mpinv{\alpha_\mathbf{Y'}}$. Therefore IC and IIL simplifies as:
\begin{eqnarray*}
\EIC{\absi,\mathbf{X'},\mathbf{Y'}} = D_{JSD}(\alpha_\mathbf{Y'} \mu, \mu' \alpha_\mathbf{X'}) = D_{JSD}(\mu, \mu' \alpha_\mathbf{X'}), \\
\EIIL{\absi,\mathbf{X'},\mathbf{Y'}} = D_{JSD}( \mu, \mpinv{\alpha_\mathbf{Y'}} \mu' \alpha_\mathbf{X'} ) =  D_{JSD}( \mu, \mu' \alpha_\mathbf{X'} )
\end{eqnarray*}
Hence, $\EIC{\absi,\mathbf{X'},\mathbf{Y'}} = \EIIL{\absi,\mathbf{X'},\mathbf{Y'}}$.

Moreover, ISIL and ISC simplifies as:
\begin{eqnarray*}
\EISIL{\absi,\mathbf{X'},\mathbf{Y'}} = D_{JSD}(\mu', \alpha_\mathbf{Y'} \mu \mpinv{\alpha_\mathbf{X'}}) = D_{JSD}(\mu', \mu \mpinv{\alpha_\mathbf{X'}}).\\
\EISC{\absi,\mathbf{X'},\mathbf{Y'}} = D_{JSD}( \mu \mpinv{\alpha_\mathbf{X'}}, \mpinv{\alpha_\mathbf{Y'}} \mu') = D_{JSD}( \mu \mpinv{\alpha_\mathbf{X'}}, \mu').
\end{eqnarray*}
By symmetry (J3), $\EISIL{\absi,\mathbf{X'},\mathbf{Y'}} = \EISC{\absi,\mathbf{X'},\mathbf{Y'}}$. $\blacksquare$

\section{Properties of the overall IC error} \label{app:overallerrorproperties}

\subsection{Property (O1): Triangle Inequality}
\begin{proposition*}
Given two abstractions $\absi$ from $\scm$ to $\scmi$, and $\absii$ from $\scmi$ to $\scmii$, $\eI{\absii\absi} \leq \eI{\absi} + \eI{\absii}$.
\end{proposition*}

\emph{Proof.} Let $\eI{\absi} = k_1 >0$. This means that:
\begin{equation*}
    \sup_{(\mathbf{X'},\mathbf{Y'}) \in \mathcal{J}} \EI{\absi,\mathbf{X'},\mathbf{Y'}} = k_1,
\end{equation*}
that is, $k_1$ is the highest interventional error for $\absi$.

Similarly, let us assume $\eI{\absii} = k_2 >0$, meaning that the highest interventional error for $\absii$ amounts to $k_2$.

By (E1), we know that the triangle equality holds for the composition of interventional error: $\EI{\absii\absi,\mathbf{X''},\mathbf{Y''}} \leq \EI{\absi,\mathbf{X'},\mathbf{Y'}} + \EI{\absii,\mathbf{X''},\mathbf{Y''}}$. In the worst case scenario we then have:
\begin{equation*}
    \EI{\absii\absi,\mathbf{X''},\mathbf{Y''}} \leq k_1 + k_2,
\end{equation*}
that is:
\begin{equation*}
    \EI{\absii\absi,\mathbf{X''},\mathbf{Y''}} \leq \eI{\absi} + \eI{\absii}.
\end{equation*}
As the worst case scenario corresponds to the supremum
\begin{equation*}
    \sup_{(\mathbf{X'},\mathbf{Y'}) \in \mathcal{K''}} \EI{\absii\absi,\mathbf{X''},\mathbf{Y''}},
\end{equation*}
by definition, we have:
$
    \eI{\absii\absi} \leq \eI{\absi} + \eI{\absii}
$. $\blacksquare$

\subsection{Property (O2): Non-monotonicity}
\begin{proposition*}
Given two abstractions $\absi$ from $\scm$ to $\scmi$, and $\absii$ from $\scmi$ to $\scmii$, it is not guaranteed that $\eI{\absii\absi} \geq \eI{\absi}$.
\end{proposition*}

\emph{Proof.} We will prove this statement by reducing this proof to (E2).
Let us consider the models $\scm$, $\scmi$ and $\scmii$, each one made up by two binary variables ($\scm[\mathbf{X}], \scmi[\mathbf{X'}], \scmii[\mathbf{X''}], \scm[\mathbf{Y}], \scmi[\mathbf{Y'}], \scmii[\mathbf{Y''}]$) and stochastic matrices ($\mu,\mu',\mu''$) as defined in the proof to (E2). Further, let us consider an abstraction between them made up of identity matrices, as in the proof to (E2).

It follows that the assessment sets $\mathcal{J}$ and $\mathcal{J'}$ for $\absi$ and $\absii$ respectively, contain only one element. Therefore, $\EI{\absi,\mathbf{X'},\mathbf{Y'}}=\eI{\absi}$ and $\EI{\absii,\mathbf{X''},\mathbf{Y''}}=\eI{\absii}$. Similarly, $\EI{\absii\absi,\mathbf{X''},\mathbf{Y''}}=\eI{\absii\absi}$.

It is then immediate that $\EI{\absii\absi,\mathbf{X''},\mathbf{Y''}} < \EI{\absi,\mathbf{X'},\mathbf{Y'}}$ translates into $\eI{\absii\absi} < \eI{\absi}$, meaning that it is it is not guaranteed that $\eI{\absii\absi} \geq \eI{\absi}$. $\blacksquare$

\subsection{Property (O3): Zeros at identity}
\begin{proposition*}
Given an abstraction $\absi$ from $\scm$ to $\scmi$, if $\absi$ is an identity, then $\eI{\absi}=0$.
\end{proposition*}

Trivially, if the abstraction is an identity, all interventional errors $\EI{\absi,\mathbf{X'},\mathbf{Y'}}$ reduce to zero by (E3). Hence, $\eI{\absi}=0$. $\blacksquare$

\subsection{Property (O4): Zeros at singleton}
\begin{proposition*}
Given an abstraction $\absi$ from $\scm$ to $\scmi$, if $\scmi$ is a singleton, then $\eIC{\absi}=0$.
\end{proposition*}

\emph{Proof.} Trivially, if $\scmi$ is a singleton, then the diagram of every error $\EIC{\absi,\mathbf{X'},\mathbf{Y'}}$ is reduced as in (E4), and $\EIC{\absi,\mathbf{X'},\mathbf{Y'}}=0$. Hence, $\eIC{\absi}=0$. $\blacksquare$

\subsection{Extension of Proposition 1}
\begin{proposition*}[Relationship between measures]
We have a partial ordering among the interventional measures as:
$\eIIL{\absi} \geq \eIC{\absi}$, 
$\eIIL{\absi} \geq \eISC{\absi}$,
$\eIC{\absi} \geq \eISIL{\absi}$,
$\eISC{\absi} \geq \eISIL{\absi}$.
\end{proposition*}

\emph{Proof.} This ordering immediately follow from Proposition 1 through the supremum.

Let $e_1({\absi}) \geq e_2({\absi})$ be one of the above relations, for which it holds by Proposition 1 that $E_1({\absi,\mathbf{X'},\mathbf{Y'}}) \geq E_2({\absi,\mathbf{X'},\mathbf{Y'}})$.

Let $e_2({\absi})=k_2$. This means that:
\begin{equation*}
    \sup_{(\mathbf{X'},\mathbf{Y'}) \in \mathcal{J}} E_2({\absi,\mathbf{X'},\mathbf{Y'}}) = k_2,
\end{equation*}
that is, $k_2$ is the highest interventional error for a pair $(\mathbf{X'},\mathbf{Y'}) \in \mathcal{J}$.

Now, by Proposition 1, it holds that, for the same pair $(\mathbf{X'},\mathbf{Y'})$, $E_1({\absi,\mathbf{X'},\mathbf{Y'}}) \geq E_2({\absi,\mathbf{X'},\mathbf{Y'}})$. Hence, $e_1({\absi}) \geq e_2({\absi})$. $\blacksquare$

\subsection{Proposition 2}
\begin{proposition}[Finiteness of the overall error] Given an abstraction $\absi$ from  $\scm$ to $\scmi$, $\eI{\absi} < \infty$ if $a$ is order-preserving.
\end{proposition}

\emph{Proof.} If $a$ is not order-preserving, then there exist $\mathbf{X},\mathbf{Y}$ in the base SCM $\scm$, such that $\mathbf{X} \prec \mathbf{Y}$ and $\mathbf{Y'} \prec \mathbf{X'}$ in the abstracted SCM $\scm'$, where $\mathbf{X'} = a(\mathbf{X})$ and $\mathbf{Y'}=a(\mathbf{Y})$. 

In evaluating any interventional measure of abstraction approximation, this would give rise to the following diagram:

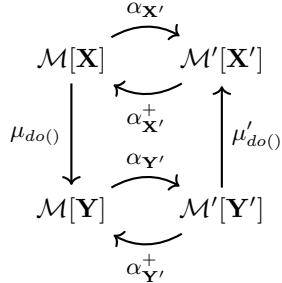
\begin{center}
\begin{tikzpicture}[shorten >=1pt, auto, node distance=1cm, thick, scale=1.0, every node/.style={scale=1.0}]
\tikzstyle{node_style} = []

\node[node_style] (M0_0) at (0,0) {$\scm[\mathbf{X}]$};
\node[node_style] (M0_1) at (2,0) {$\scmi[\mathbf{X'}]$};
\node[node_style] (M1_0) at (0,-2) {$\scm[\mathbf{Y}]$};
\node[node_style] (M1_1) at (2,-2) {$\scmi[\mathbf{Y'}]$};

\draw[->,bend left]  (M0_0) to node[above,font=\small]{$\alpha_\mathbf{X'}$} (M0_1);
\draw[->,bend left]  (M1_0) to node[above,font=\small]{$\alpha_\mathbf{Y'}$} (M1_1);
\draw[->,bend left]  (M0_1) to node[below,font=\small]{$\mpinv{\alpha_\mathbf{X'}}$} (M0_0);
\draw[->,bend left]  (M1_1) to node[below,font=\small]{$\mpinv{\alpha_\mathbf{Y'}}$} (M1_0);

\draw[->]  (M0_0) to node[left,font=\small]{$\mu_{do()}$} (M1_0);
\draw[->]  (M1_1) to node[right,font=\small]{$\mu'_{do()}$} (M0_1);
\end{tikzpicture}
\end{center}

Because of the inverted directionality of $\mu'_{do()}$ no interventional measure (IC, IIL, ISC, ISIL) can be computed by definition.

Moreover, from a causal point of view, interventions on the base model act on causes ($\mathbf{X}$) while the equivalent intervention on the abstracted model would act on the effect ($\mathbf{X'}$). 

This points to an underlying incompatibility between the two models. Hence, any measure of abstraction approximation would equal $\infty$. $\blacksquare$

\section{Algorithms} \label{app:algorithms}

\subsection{Abstraction Learning}

Alg. \ref{alg:eIlearner} presents the pseudo-code for learning an optimal abstraction by enumeration. A discussion of this algorithm is available in the main text in Sec. \ref{sec:evaluating}.

\begin{algorithm}[tb]
    \caption{Abstraction learning}
    \label{alg:eIlearner}
    \textbf{Input}: $\scm$, $\scmi$, ${I} \in \{IC,IIL,ISIL,ISC\}$, $\mathcal{J}$, $\mathcal{K}$\\
    \textbf{Output}: optimal $\absi$
    \begin{algorithmic}[1] 
        \IF[$O(|E|)$] {$a$ is not order-preserving}
            \STATE \textbf{return} $\mathbf{None}$
        \ENDIF
        \STATE Initialize $E_{opt}=\infty, \alpha_{opt} = \mathbf{None}$

        \FOR[$O(|\mathcal{K}|)$]{$\absi \in \mathcal{K}$}
            \STATE $\eI{\absi} =$ Alg.\ref{alg:eIevaluator}($\scm$, $\scmi$, $\absi$, {I}, $\mathcal{J}$) \COMMENT{$O(|\mathcal{J}|)$}
            \IF {$\eI{\absi} < E_{opt}$}
                \STATE $\alpha_{opt} \leftarrow \absi$
                \STATE $E_{opt} \leftarrow \eI{\absi}$
            \ENDIF
        \ENDFOR
        \STATE \textbf{return} $\alpha_{opt}$
    \end{algorithmic}
\end{algorithm}

\section{Sample Models and Abstractions} \label{app:samplemodels}
Here we provide a full specifications of the models and abstractions used in the examples throughout the paper.

\subsection{Model in Figure \ref{fig:SCM0}}
The SCM for the model of Fig. \ref{fig:SCM0}, as described in Ex. \ref{ex:SCM0SCM1}, is defined as follows:
\begin{center}
\begin{tabular}{c}
$\mathcal{X}=\{Sm,Tar,LC\}$
\tabularnewline
$\scm[Sm] = \{0,1\}$
\tabularnewline
$\scm[Tar] = \{0,1\}$
\tabularnewline
$\scm[LC] = \{0,1\}$
\tabularnewline
\end{tabular}
\par\end{center}

The stochastic matrices for the mechanisms are defined as follows:
\[
\mu_{\cdot \rightarrow Sm}\left[\begin{array}{c}
0.8 \\
0.2
\end{array}\right],
\]
\[
\mu_{Sm \rightarrow Tar}\left[\begin{array}{cc}
1.0 & 0.2\\
0.0 & 0.8
\end{array}\right],
\]
\[
\mu_{Tar \rightarrow LC}\left[\begin{array}{cc}
0.9 & 0.6\\
0.1 & 0.4
\end{array}\right].
\]

\subsection{Model in Figure \ref{fig:SCM1}} \label{app:SCM1}
The SCM for the model of Fig. \ref{fig:SCM1}, as described in Ex. \ref{ex:SCM0SCM1}, is defined as follows:
\begin{center}
\begin{tabular}{c}
$\mathcal{X'}=\{Sm',Hea'\}$
\tabularnewline
$\scmi[Sm'] = \{0,1\}$
\tabularnewline
$\scmi[Hea'] = \{0,1\}$
\tabularnewline
\end{tabular}
\par\end{center}

The stochastic matrices for the mechanisms are defined as follows:
\[
\nu_{\cdot \rightarrow Sm'}\left[\begin{array}{c}
0.5 \\
0.5
\end{array}\right],
\]
\[
\nu_{Sm' \rightarrow Hea'}\left[\begin{array}{cc}
0.4 & 0.2\\
0.6 & 0.8
\end{array}\right].
\]

\subsection{Model in Figure \ref{fig:SCM2}} \label{app:SCM2}
The SCM for the model of Fig. \ref{fig:SCM2} is defined as follows:
\begin{center}
\begin{tabular}{c}
$\mathcal{X'}=\{Env'',Gen'',LC''\}$
\tabularnewline
$\scmii[Env''] = \{0,1,2\}$
\tabularnewline
$\scmii[Gen''] = \{0,1\}$
\tabularnewline
$\scmii[LC''] = \{0,1\}$
\tabularnewline
\end{tabular}
\par\end{center}

The stochastic matrices for the mechanisms are defined as follows:
\[
\nu_{\cdot \rightarrow Env''}\left[\begin{array}{c}
0.7 \\
0.1 \\
0.2
\end{array}\right],
\]
\[
\nu_{\cdot \rightarrow Gen''}\left[\begin{array}{c}
0.3 \\
0.7
\end{array}\right],
\]
\[
\nu_{Env'' \times Gen'' \rightarrow LC''}\left[\begin{array}{cccccc}
	0.7 & 0.6 & 0.5 & 0.4 & 0.4 & 0.3\\
	0.3 & 0.4 & 0.5 & 0.6 & 0.6 & 0.7
\end{array}\right].
\]

\subsection{Model in Figure \ref{fig:SCMstar}}
The SCM for the model of Fig. \ref{fig:SCMstar} is defined as follows:
\begin{center}
\begin{tabular}{c}
$\mathcal{X}=\{*\}$
\tabularnewline
$\scm[*] = \{*\}$
\tabularnewline
\end{tabular}
\par\end{center}

The stochastic matrix for the only mechanism is defined as follows:
\[
\nu_{* \rightarrow *}\left[\begin{array}{c}
1.0 
\end{array}\right].
\]

\subsection{Abstraction from Model of Figure \ref{fig:SCM0} to Model of Figure \ref{fig:SCM1}}

The abstraction $\absi$ from the model $\scm$ of Fig. \ref{fig:SCM0} to the model $\scmi$ of Fig. \ref{fig:SCM1}, as described in Ex. \ref{ex:SCM0SCM1}, is defined by:
\begin{center}
	\begin{tabular}{c}
		$R=\{Sm, LC\}$
		\tabularnewline
	\end{tabular}
	\par\end{center}
\begin{center}
	\begin{tabular}{cc}
		$a:$ & $ \{ Sm \mapsto Sm',$
		\tabularnewline
		& $LC \mapsto Hea' \}$
		\tabularnewline
	\end{tabular}
\par\end{center}

\begin{center}
	\begin{tabular}{cc}
		$alpha_{Sm'}:$ & $ \left[\begin{array}{cc}
            1 & 0\\
            0 & 1
            \end{array}\right],$
		\tabularnewline
		$alpha_{Hea'}:$ & $ \left[\begin{array}{cc}
            1 & 0\\
            0 & 1
            \end{array}\right].$
		\tabularnewline
	\end{tabular}
\par\end{center}

If we evaluate the overall IC error wrt the assessment set $\mathcal{J} = \{ (Sm',Hea') \}$ as in Ex. \ref{ex:SCM0SCM1_abs}, we obtain the following result:

\begin{align*}
\eIC{\absi} & = \sup_{(\mathbf{X'},\mathbf{Y'}) \in \mathcal{J}} \EIC{\absi,\mathbf{X'},\mathbf{Y'}}\\
 & = \EIC{\absi,Sm',Hea'}\\
 & = D_{JSD} (\alpha_{Hea'} \mu_{Tar\rightarrow LC} \mu_{Sm\rightarrow Tar}, \nu_{Sm' \rightarrow Hea'} \alpha_{Sm'})\\
 & = D_{JSD} (\left[\begin{array}{cc}
            1 & 0\\
            0 & 1
            \end{array}\right] \left[\begin{array}{cc}
            0.9 & 0.6\\
            0.1 & 0.4
            \end{array}\right] \left[\begin{array}{cc}
            1.0 & 0.2\\
            0.0 & 0.8
            \end{array}\right],\\
            & \qquad \left[\begin{array}{cc}
            0.4 & 0.2\\
            0.6 & 0.8
            \end{array}\right] \left[\begin{array}{cc}
            1 & 0\\
            0 & 1
            \end{array}\right])\\
& = D_{JSD} (\left[\begin{array}{cc}
            0.9 & 0.66\\
            0.1 & 0.34
            \end{array}\right], \left[\begin{array}{cc}
            0.4 & 0.2\\
            0.6 & 0.8
            \end{array}\right])\\
& \approx 0.385 \\
\end{align*}

\section{Experimental Details} \label{app:experimentaldetails}

Here we provide all the details about our simulations. Implementation of all the simulations is available online\footnote{\url{https://github.com/FMZennaro/CausalAbstraction/tree/main/papers/2023-quantifying-consistency-and-infoloss}}.

\subsection{LUCAS model}

\begin{figure}
    \centering
		\begin{tikzpicture}[shorten >=1pt, auto, node distance=1cm, thick, scale=0.7, every node/.style={scale=0.7}]
			\tikzstyle{node_style} = [circle,draw=black]
			\node[node_style] (Anx) at (0,0) {Anx};
                \node[node_style] (PP) at (2,0) {PP};
			\node[node_style] (BED) at (4,0) {BED};

                \node[node_style] (YF) at (-1,-2) {YF};
                \node[node_style] (Smo) at (1,-2) {Sm};
			\node[node_style] (Gen) at (3,-2) {Gen};

                \node[node_style] (All) at (0,-4) {All};
                \node[node_style] (LC) at (2,-4) {LC};
			\node[node_style] (AD) at (4,-4) {AD};

                \node[node_style] (Cou) at (-1,-6) {Cou};
                \node[node_style] (Fat) at (1,-6) {Fat};
			\node[node_style] (CA) at (3,-6) {CA};
			
			\draw[->]  (Anx) to (Smo);
                \draw[->]  (PP) to (Smo);
                \draw[->]  (Smo) to (YF);
                \draw[->]  (Smo) to (LC);
                \draw[->]  (Gen) to (AD);
                \draw[->]  (Gen) to (LC);
                \draw[->]  (LC) to (Cou);
                \draw[->]  (LC) to (Fat);
                \draw[->]  (All) to (Cou);
                \draw[->]  (Cou) to (Fat);
                \draw[->]  (AD) to (CA);
                \draw[->]  (Fat) to (CA);
		\end{tikzpicture}
		\caption{DAG underlying the LUCAS model, reproduced from \url{http://www.causality.inf.ethz.ch/data/LUCAS.html}}
		\label{fig:LUCAS0}
\end{figure}
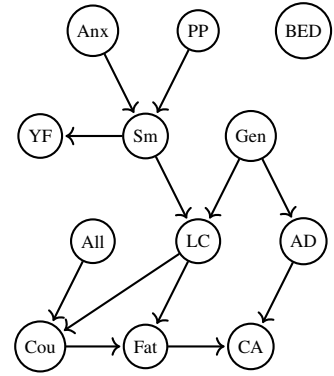

For our base model, we use the lung cancer toy model from \url{http://www.causality.inf.ethz.ch/data/LUCAS.html}. Fig. \ref{fig:LUCAS0} shows the DAG underlying the base model $\scm$; Tab. \ref{tab:variablenames} reports the acronyms and the name of the variables.
The SCM $\scm$ is defined over the set of variables $\mathcal{X}$ comprising all the endogenous nodes in Fig. \ref{fig:LUCAS0}; the stochastic matrices $\mu$ between the nodes take the values originally defined in \url{http://www.causality.inf.ethz.ch/data/LUCAS.html}.

\begin{table}
\begin{centering}
\begin{tabular}{cc}
\hline
\emph{Anx} & Anxiety\tabularnewline
\emph{PP} & Peer Pressure\tabularnewline
\emph{Sm} & Smoking\tabularnewline
\emph{YF} & Yellow Fingers\tabularnewline
\emph{BED} & Born on Even Day\tabularnewline
\emph{Gen} & Genetics\tabularnewline
\emph{LC} & Lung Cancer\tabularnewline
\emph{All} & Allergy\tabularnewline
\emph{AD} & Attention Disorder\tabularnewline
\emph{Cou} & Coughing\tabularnewline
\emph{Fat} & Fatigue\tabularnewline
\emph{CA} & Car Accident\tabularnewline
\emph{Hea} & Health\tabularnewline
\emph{Env} & Environment\tabularnewline
\hline
\end{tabular}
\par\end{centering}
\caption{Acronyms of the variables.}\label{tab:variablenames}
\end{table}

\subsection{Health Scenario} \label{app:heathscenario}

In the first simulation, Lab A works with the LUCAS model defined above, while Lab B relies on the simpler model defined below.

\paragraph{Health SCM.}
The SCM $\scmi$ of Lab B is made up by two binary variables meant to capture the effect of smoking on health (see Fig. \ref{fig:SCM1}). The SCM is defined as the model in App. \ref{app:SCM1} with the exception for the stochastic matrix for the mechanism $Sm' \rightarrow Hea'$ which is not fully defined, and for which the agent has considered three alternatives:
\[
\mathcal{K}=\left\{ \left[\begin{array}{cc}
0.3 & 0.2\\
0.7 & 0.8
\end{array}\right],\left[\begin{array}{cc}
0.4 & 0.2\\
0.6 & 0.8
\end{array}\right],\left[\begin{array}{cc}
0.5 & 0.2\\
0.5 & 0.8
\end{array}\right]\right\}.
\]

\paragraph{Abstraction.}
An abstraction $\absi$ from $\scm$ to $\scmi$ has been defined as:
\begin{center}
\begin{tabular}{c}
$R=\{Sm, Cou, Fat\}$
\tabularnewline
\end{tabular}
\par\end{center}
\begin{center}
\begin{tabular}{cc}
$a:$ & $ \{ Sm \mapsto Sm',$
\tabularnewline
& $Cou \mapsto Hea', $
\tabularnewline
& $Fat \mapsto Hea'\}$
\tabularnewline
\end{tabular}
\par\end{center}

\begin{figure}
    \centering
		\begin{tikzpicture}[shorten >=1pt, auto, node distance=1cm, thick, scale=0.7, every node/.style={scale=0.7}]
			\tikzstyle{node_style} = [circle,draw=black]
			\node[node_style] (Anx) at (0,0) {Anx};
                \node[node_style] (PP) at (2,0) {PP};
			\node[node_style] (BED) at (4,0) {BED};

                \node[node_style] (YF) at (-1,-2) {YF};
                \node[node_style] (Smo) at (1,-2) {Sm};
			\node[node_style] (Gen) at (3,-2) {Gen};

                \node[node_style] (All) at (0,-4) {All};
                \node[node_style] (LC) at (2,-4) {LC};
			\node[node_style] (AD) at (4,-4) {AD};

                \node[node_style] (Cou) at (-1,-6) {Cou};
                \node[node_style] (Fat) at (1,-6) {Fat};
			\node[node_style] (CA) at (3,-6) {CA};

                \node[node_style,blue] (Smo_) at (.5,-10) {Sm'};
                \node[node_style,blue] (Hea_) at (2.5,-10) {Hea'};
   
			\draw[->]  (Anx) to (Smo);
                \draw[->]  (PP) to (Smo);
                \draw[->]  (Smo) to (YF);
                \draw[->]  (Smo) to (LC);
                \draw[->]  (Gen) to (AD);
                \draw[->]  (Gen) to (LC);
                \draw[->]  (LC) to (Cou);
                \draw[->]  (LC) to (Fat);
                \draw[->]  (All) to (Cou);
                \draw[->]  (Cou) to (Fat);
                \draw[->]  (AD) to (CA);
                \draw[->]  (Fat) to (CA);

                \draw[->,blue]  (Smo_) to (Hea_);

                \draw[->,blue,dashed,bend right=50]  (Smo) to (Smo_);  
                \draw[->,blue,dashed]  (Fat) to (Hea_);
                \draw[->,blue,dashed]  (Cou) to (Hea_);
                
		\end{tikzpicture}
		\caption{Abstraction (blue dashed arrow) from the base model $\scm$ (black) to the abstracted model $\scmi$ (blue) in the health exam scenario.}
		\label{fig:Hscenario}
\end{figure}
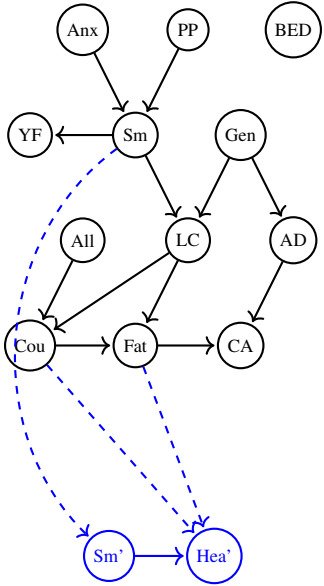

See Fig. \ref{fig:Hscenario} for reference. In this abstraction, both $\scmi$ and $\absi$ are only partially specified: for $\scmi$, there is a small set $\mathcal{K}$ of three possible stochastic matrices $\nu$ connecting $Sm'$ to $Hea'$; for $\absi$, there is no specification at all for $\alpha_{Sm'}$ and $\alpha_{Hea'}$.

\paragraph{Abstraction Learning.} We perform abstraction learning using Alg. \ref{alg:eIlearner}, and iterating over $\mathcal{K}$ and over the set of all possible abstraction matrices $\alpha_{Sm'}$ and $\alpha_{Hea'}$. We use the causal assessment set $\mathcal{J}_c$ as an assessment set. Since we are learning by considering interventions from the perspective of Lab A, we compute IC and IIL.

Optimal IC error $\eIC{\absi} = 0.029$ is achieved for:
$$
\nu = \left[\begin{array}{cc}
0.4 & 0.2\\
0.6 & 0.8
\end{array}\right], \alpha_{Sm'} \left[\begin{array}{cc}
1 & 0\\
0 & 1
\end{array}\right], \alpha_{Hea'} \left[\begin{array}{cccc}
0 & 1 & 1 & 0\\
1 & 0 & 0 & 1
\end{array}\right].
$$

Optimal IIL error $\eIIL{\absi} = 0.160$ is achieved for:
$$
\nu = \left[\begin{array}{cc}
0.5 & 0.2\\
0.5 & 0.8
\end{array}\right], \alpha_{Sm'} \left[\begin{array}{cc}
1 & 0\\
0 & 1
\end{array}\right], \alpha_{Hea'} \left[\begin{array}{cccc}
1 & 1 & 1 & 0\\
0 & 0 & 0 & 1
\end{array}\right].
$$

Meaning and implication of these results are discussed in the main text.

\paragraph{Downstream task: exam}

In the first downstream task, Lab A and Lab B perform experiments on patients, compute a health index, and provide this result to a decision-making module that will establish whether the patient requires further examination. The decision-making module has been trained on data produced by the model of Lab A. For the sake of accuracy and fairness, we would like the distribution produced by the abstracted model of Lab B to be close to the one produced by Lab A.

To evaluate the closeness of the results, we empirically compare the distribution produced by the model of Lab A and Lab B after abstraction learning. We take as a reference distribution $\hat{P}(\alpha_{Hea'}(Hea) \vert do(Sm))$; this defines the distribution of the health index (computed by aggregating the base-model variables $Cou,Fat$ via $\alpha_{Hea'}$) when Lab A performs an intervention. We compare this distribution to the distribution $\hat{P}(Hea'\vert do(\alpha_{Sm'}(Sm)))$, which accounts for the health index in the abstracted model when Lab B performs an intervention. The two distributions are evaluated when using an abstraction that optimizes IC or IIL. Results are reported in Tab. \ref{tab:health1}.

\paragraph{Downstream task: car accident}

In the second downstream task, Lab A relies on the abstraction provided by Lab B in order to simplify the computational load of its own model. Specifically, given an experiment affecting the variable $Sm$, Lab A wants to compute the value of variables $Fat$ and $Cou$ using the lighter model of Lab B, and then estimate the probability of car accident $CA$. For the sake of accuracy, we would like the distribution for car accident in the original model of Lab A and the distribution in the model where we relied on the abstraction of Lab B to be as close as possible.

To evaluate the closeness of the results, we empirically compare the distribution produced by the model of Lab A and Lab B after abstraction learning. We take as a reference distribution $\hat{P}(CA \vert do(Sm))$; this defines the distribution of car accidents when computing directly with the original model of Lab A. We compare this distribution to the distribution $\hat{P}_I(CA \vert do(Sm))$, which evaluates the same quantity but relying on the model of Lab B to compute the value of $Cou$ and $Fat$ in the base model, before estimating $CA$. Distributions are evaluated when using an abstraction that optimizes IC or IIL. Results are reported in Tab. \ref{tab:health2}.

\subsection{Lung Cancer Scenario} \label{app:lungcancerscenario}

In the second simulation, Lab A works again with the LUCAS model, while Lab C now relies on a different high-level model defined below.

\paragraph{Lung Cancer SCM.}
The SCM $\scmii$ of Lab C is made up by three variables meant to capture the effect of environmental and genetic factors on lung cancer (see Fig. \ref{fig:SCM2}). The SCM is defined as the model in App. \ref{app:SCM2}.

\paragraph{Abstraction.}
An abstraction $\absi$ from $\scm$ to $\scmii$ has been defined as:
\begin{center}
	\begin{tabular}{c}
		$R=\{Anx, PP, Gen, All, LC\}$
		\tabularnewline
	\end{tabular}
	\par\end{center}
\begin{center}
	\begin{tabular}{cc}
		$a:$ & $ \{ Anx \mapsto Env'',$
		\tabularnewline
		& $PP \mapsto Env'', $
		\tabularnewline
		& $Gen \mapsto Gen'',$
		\tabularnewline
		& $All \mapsto Gen'',$
		\tabularnewline
		& $LC \mapsto LC''\}$
		\tabularnewline
	\end{tabular}
	\par\end{center}

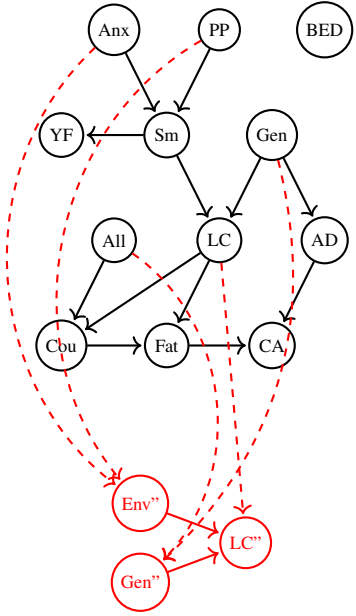
\begin{figure}
	\centering
	\begin{tikzpicture}[shorten >=1pt, auto, node distance=1cm, thick, scale=0.7, every node/.style={scale=0.7}]
		\tikzstyle{node_style} = [circle,draw=black]
		\node[node_style] (Anx) at (0,0) {Anx};
		\node[node_style] (PP) at (2,0) {PP};
		\node[node_style] (BED) at (4,0) {BED};
		
		\node[node_style] (YF) at (-1,-2) {YF};
		\node[node_style] (Smo) at (1,-2) {Sm};
		\node[node_style] (Gen) at (3,-2) {Gen};
		
		\node[node_style] (All) at (0,-4) {All};
		\node[node_style] (LC) at (2,-4) {LC};
		\node[node_style] (AD) at (4,-4) {AD};
		
		\node[node_style] (Cou) at (-1,-6) {Cou};
		\node[node_style] (Fat) at (1,-6) {Fat};
		\node[node_style] (CA) at (3,-6) {CA};
		
		\node[node_style,red] (Env_) at (.5,-9) {Env''};
		\node[node_style,red] (Gen_) at (.5,-10.5) {Gen''};
		\node[node_style,red] (LC_) at (2.5,-9.75) {LC''};
		
		\draw[->]  (Anx) to (Smo);
		\draw[->]  (PP) to (Smo);
		\draw[->]  (Smo) to (YF);
		\draw[->]  (Smo) to (LC);
		\draw[->]  (Gen) to (AD);
		\draw[->]  (Gen) to (LC);
		\draw[->]  (LC) to (Cou);
		\draw[->]  (LC) to (Fat);
		\draw[->]  (All) to (Cou);
		\draw[->]  (Cou) to (Fat);
		\draw[->]  (AD) to (CA);
		\draw[->]  (Fat) to (CA);
		
		\draw[->,red]  (Env_) to (LC_);
		\draw[->,red]  (Gen_) to (LC_);
		
		\draw[->,red,dashed,bend right=50]  (PP) to (Env_);
		\draw[->,red,dashed,bend right=50]  (Anx) to (Env_);
		\draw[->,red,dashed]  (LC) to (LC_);
		\draw[->,red,dashed,bend left=50]  (All) to (Gen_);
		\draw[->,red,dashed,bend left=30]  (Gen) to (Gen_);
		
	\end{tikzpicture}
	\caption{Abstraction (red dashed arrow) from the base model $\scm$ (black) to the abstracted model $\scmii$ (red) in the lung cancer scenario.}
	\label{fig:LCscenario}
\end{figure}

See Fig. \ref{fig:LCscenario} for reference. Abstraction maps $\alpha_{Env''}$, $\alpha_{Gen''}$, and $\alpha_{LC''}$ are not specified.

\paragraph{Abstraction Evaluation.} We perform abstraction evaluation using Alg. \ref{alg:eIevaluator}, and iterating over the set of all possible abstraction matrices $\alpha_{Env''}$, $\alpha_{Gen''}$, and $\alpha_{LC''}$. Since we are learning considering interventions from the perspective of Lab C, we compute ISIL. We consider three possible assessment sets: causal  $\mathcal{J}_c = \{ (\{Env''\},\{LC''\}), (\{Gen''\},\{LC''\}), (\{Env'',Gen''\},\{LC''\})\}$, parental $\mathcal{J}_p = \{(\{Env'',Gen''\},\{LC''\})\}$, and custom $\mathcal{J}_u = \{ (\{Env''\},\{LC''\})\}$.

Optimal ISIL when using $\mathcal{J}_c$ is achieved at $\eISIL{\absi} = 0.254$ for:
$$
\alpha_{Env''} \left[\begin{array}{cccc}
	0 & 0 & 1 & 1\\
	0 & 1 & 0 & 0\\
        1 & 0 & 0 & 0
\end{array}\right], \alpha_{Gen''} \left[\begin{array}{cccc}
	0 & 1 & 1 & 1\\
	1 & 0 & 0 & 0
\end{array}\right], \alpha_{LC''} \left[\begin{array}{cc}
	0 & 1\\
	1 & 0
\end{array}\right].
$$

Optimal ISIL when using $\mathcal{J}_p$ is achieved at $\eISIL{\absi} = 0.221$ for:
$$
\alpha_{Env''} \left[\begin{array}{cccc}
	1 & 0 & 0 & 0\\
	0 & 1 & 0 & 0\\
        0 & 0 & 1 & 1
\end{array}\right], \alpha_{Gen''} \left[\begin{array}{cccc}
	1 & 0 & 0 & 0\\
	0 & 1 & 1 & 1
\end{array}\right], \alpha_{LC''} \left[\begin{array}{cc}
	1 & 0\\
	0 & 1
\end{array}\right].
$$

Optimal ISIL when using $\mathcal{J}_u$ is achieved at $\eISIL{\absi} = 0.129$ for:
$$
\alpha_{Env''} \left[\begin{array}{cccc}
	1 & 0 & 0 & 0\\
	0 & 1 & 0 & 0\\
        0 & 0 & 1 & 1
\end{array}\right], \alpha_{Gen''} \left[\begin{array}{cccc}
	1 & 0 & 0 & 0\\
	0 & 1 & 1 & 1
\end{array}\right], \alpha_{LC''} \left[\begin{array}{cc}
	1 & 0\\
	0 & 1
\end{array}\right].
$$

Notice that, although abstraction evaluation with $\mathcal{J}_p$ and $\mathcal{J}_u$ achieve a different ISIL value, they still reach this minimum wrt the same abstraction matrices.

\paragraph{Downstream task: lung cancer prediction}

In the downstream task, Lab A and Lab C perform experiments on patients, and provide estimates about lung cancer probability to a decision-making module trained on data produced by the model of Lab C. For the sake of accuracy and fairness, we would like the distribution produced by the base model of Lab A to be close to the one produced by Lab C.

To evaluate the closeness of the results, we empirically compare the distribution produced by the model of Lab C and Lab A after abstraction learning. We take as a reference distribution $\hat{P}(LC'' \vert do(Env''))$; this defines the distribution of lung cancer when Lab C performs an intervention. We compare this distribution to the distribution $\hat{P}_{\mathcal{J}}(\alpha_{LC''}(LC)\vert do(\mpinv{\alpha_{Env''}}(Env'')))$, which accounts for the distribution of lung cancer in the abstracted model when Lab A performs an intervention. The two distributions are optimized when using an abstraction evaluated wrt different assessment sets. Results are reported in Tab. \ref{tab:lc1}. Tab. \ref{tab:lc1-time} reports the wall-clock times required to run Alg. \ref{alg:eIlearner} in order to compute the ISIL-minimizing abstraction.

\begin{table}
	\resizebox{\columnwidth}{!}{%
		\begin{centering}
			\begin{tabular}{cc}
				\hline 
				& Time(s)\tabularnewline
				\hline
                    $\hat{P}(LC''=1 \vert do(Env''))$  & -\tabularnewline
				$\hat{P}_{\mathcal{J}_{c}}(\alpha_{LC''}(LC=1) \vert \mpinv{\alpha_{Env''}}( do(Env'')))$ &  23.73 $\pm$ 0.04 \tabularnewline
				$\hat{P}_{\mathcal{J}_{p}}(\alpha_{LC''}(LC=1) \vert \mpinv{\alpha_{Env''}}( do(Env'')))$ &  8.79 $\pm$ 0.06\tabularnewline
				$\hat{P}_{\mathcal{J}_{u}}(\alpha_{LC''}(LC=1) \vert \mpinv{\alpha_{Env''}}( do(Env'')))$ &  8.01 $\pm$ 0.05  \tabularnewline
				\hline 
			\end{tabular}
			\par\end{centering}}
	\caption{Comparison of wall-clock time required to evaluate the best ISIL-minimizing abstraction wrt causal set ($\hat{P}_{\mathcal{J}_c}$), parental set ($\hat{P}_{\mathcal{J}_p}$), or custom set ($\hat{P}_{\mathcal{J}_u}$).}\label{tab:lc1-time}
\end{table}

\end{document}